\documentclass{article}

\usepackage[final]{corl_2026} 

\usepackage{graphicx}
\usepackage{xcolor}
\usepackage{wrapfig}
\usepackage{amsmath}
\usepackage{amssymb}
\usepackage{subcaption}
\usepackage{booktabs}
\usepackage{multirow}  
\usepackage{threeparttable}
\usepackage{rotating}
\usepackage{array}
\usepackage{adjustbox}
\usepackage{makecell}
\usepackage{comment}
\usepackage{algorithm}
\usepackage{algpseudocode} 
\usepackage{bm}
\usepackage{enumitem}

\newcommand{\todocomment}[1]{\textcolor{red}{TODO: #1}}

\definecolor{mitas}{RGB}{232,93,4}

\title{Multi-Resolution Tactile Imitation Learning \\ for Contact-Rich Robotic Manipulation \vspace{-0.5em}}

\author{
  Rickmer~Krohn$^{1,3,4}$, Erik~Helmut$^{2}$, Niklas~Funk$^{2}$, \\ \textbf{Jan Peters}$^{2,3,4,5}$, \textbf{Vignesh Prasad}$^{1,3,4}$ \textbf{and} \textbf{Georgia Chalvatzaki}$^{1,3,4}$ \vspace{0.5em} \\ 
  $^{1}$ Interactive Robot Perception \& Learning, TU Darmstadt\\
  $^{2}$ Intelligent Autonomous Systems, TU Darmstadt\\
    $^{3}$ Hessian AI,   $^{4}$ Robotics Institute Germany \\
    $^{5}$ German Research Center for AI (DFKI)  
    \vspace{0.5em}\\
    {\hypersetup{urlcolor=mitas}%
  \href{https://mitas-touch.github.io}{https://mitas-touch.github.io}}
}

\begin{document}
\maketitle

\vspace{-2.5em}
\begin{abstract}
Touch sensing is beneficial for solving a wide variety of manipulation tasks. While there exists a wide range of tactile sensors with different properties, exploiting the fusion of multiple heterogeneous tactile sensors to improve manipulation learning remains underexplored. We present Multi-Resolution Tactile Sensing (MiTaS), a representation framework that leverages multiple tactile sensors operating at different temporal resolutions in order to solve complex contact-rich manipulation tasks.
We propose an effective architecture using modality-specific convolutional stems and transformer-based fusion that effectively integrates information from an RGB camera stream, a vision-based GelSight Mini sensor and a high-frequency event-based Evetac sensor.
This multisensory representation then conditions a flow-matching policy for solving downstream tasks. Our systematic study across five contact-rich manipulation tasks demonstrates the effectiveness of multi-resolution tactile features in imitation learning wherein MiTaS achieves an average success rate of 80 \%, while vision-only (35.5\,\%) and visuotactile (55.5\,\%, 63\,\%) baselines cannot solve the task reliably. 


\end{abstract}





\vfill
\begin{figure}[h!]
    \centering
    \includegraphics[width=1.0\linewidth]{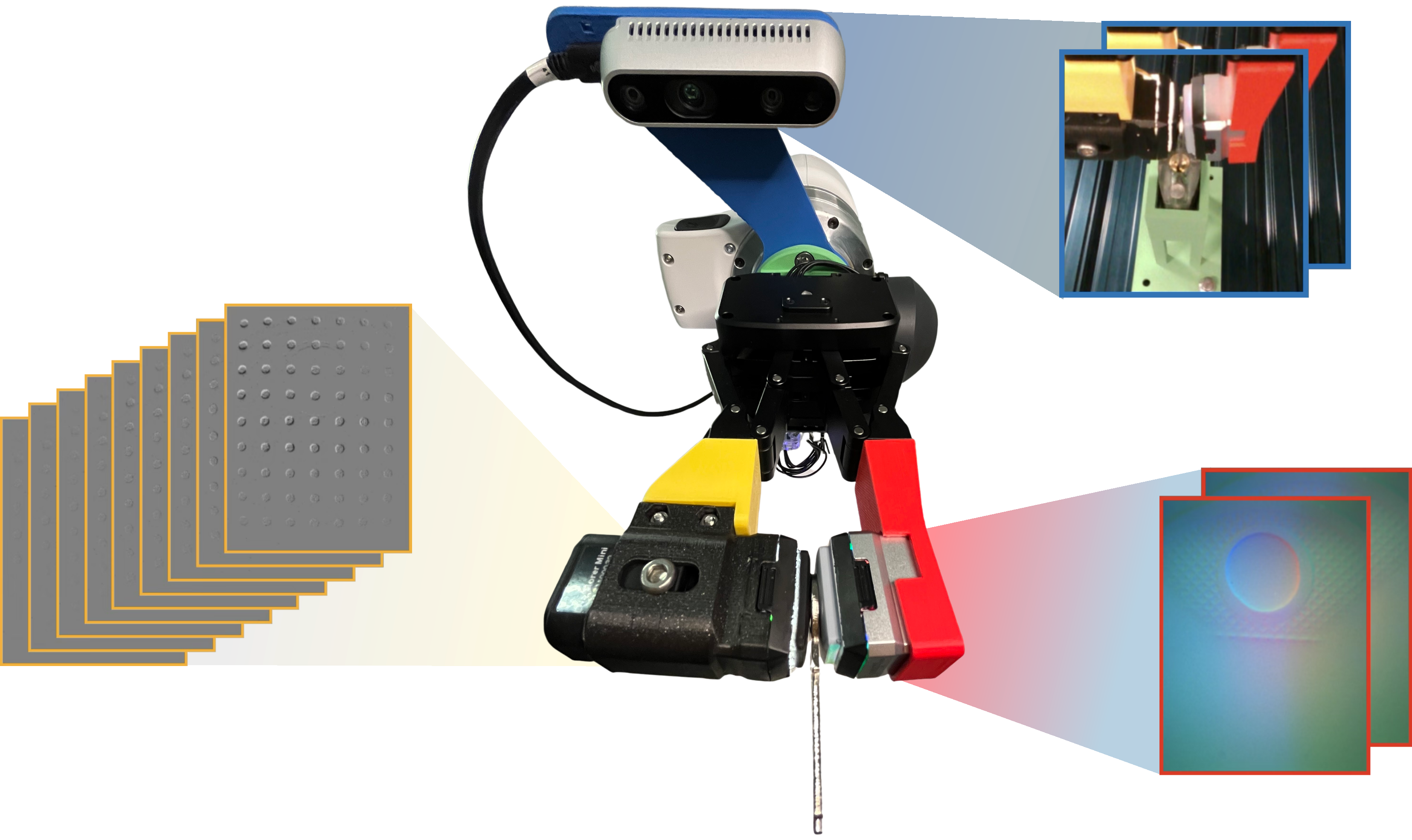}
    \caption{MiTaS combines an RGB camera (blue), a GelSight Mini (red) and an event-based Evetac sensor (yellow) mounted on the gripper. Our method fuses these multi-resolution tactile streams to condition a flow-matching policy, enabling improved control in contact-rich manipulation tasks such as key in lock.}
    \label{fig:mitas_hero}
\end{figure}


\section{Introduction}

Humans rely on multi-modal touch sensing to accomplish delicate manipulation tasks, seamlessly integrating information from diverse mechanoreceptors operating at different temporal and spatial resolutions~\cite{roland_2010}. Similarly, robotic systems can benefit from heterogeneous tactile sensing to understand objects, tasks, and contact dynamics, enabling precise manipulation capabilities.

RGB vision-based tactile sensors such as GelSight~\citep{yuan_gelsight_2017, lambeta2020digit, ward2018tactip} have shown remarkable results in manipulation by providing high-resolution spatial information about local contact geometry, deformation, and surface structure~\citep{dong2021tactile, calandra2025feelingsuccessdoestouch, helmut_tactile-conditioned_2025, hansen_visuotactile-rl_2022, yu_mimictouch_2024, yang2025anyrotate, romero2024eyesight}. However, such frame-based tactile sensing is less suited to capturing very fast contact changes, such as impact, incipient slip, high-frequency vibration. These changes may occur between frames or be smoothed out by conventional image-based processing. Recent advances in tactile sensing technology have also introduced event-based tactile cameras such as Evetac~\citep{funk_evetac_2024} or GelEvent~\cite{gelevent_2025}, which offer high temporal resolution to capture rapid contact dynamics that may otherwise go unnoticed by traditional vision-based sensors. These high-frequency signals can detect critical events such as early contact, incipient slip, vibrations during insertion, and subtle variations during sliding contact. 

Given the complementary strengths of classic vision-based and event-based sensors, a crucial question arises: can robotic systems benefit from leveraging heterogeneous tactile sensing capabilities to better understand intricate contact dynamics and improve manipulation learning?


In this work, we present \textbf{M}ult\textbf{i}-Resolution \textbf{Ta}ctile \textbf{S}ensing (MiTaS), a comprehensive representation framework to condition a flow matching imitation learning policy to leverage multiple tactile sensors, operating at different temporal resolutions, to solve complex manipulation tasks. Our key insight is that different tactile modalities capture complementary aspects of contact-rich interactions: standard vision-based sensors provide high spatial resolution of contact geometry, while event-based sensors capture high-frequency temporal dynamics. We introduce a effective architecture that extracts and fuses heterogeneous tactile features through modality-specific CNN-stems and transformer-based fusion. Additionally, we propose a multi-tactile dropout scheme that boosts the performance of a classic visuotactile policy by leveraging Evetac during training, without requiring it at inference.
 
Across five contact-rich manipulation tasks, policies using multiple heterogeneous tactile sensors consistently outperform vision-only baselines. MiTaS achieves an average success rate of 80\%, whereas vision-only policies fail to solve the tasks reliably, highlighting the benefit of combining complementary tactile sensors for both dynamic and fine-grained manipulation.
An analysis of attention scores throughout task execution reveals how sensor importance shifts during different phases of manipulation, validating our multi-resolution sensing hypothesis. To the best of our knowledge, this is the first work to combine a classic vision-based tactile sensor with an event-based tactile sensor, thereby showing how complementary tactile features enable complex contact-rich manipulation. 

Our contributions are the following: (\textbf{1}) A effective architecture for extracting and fusing heterogeneous tactile sensor readings operating at different temporal resolutions; (\textbf{2}) A multi-tactile dropout scheme that improves visuotactile policy performance on contact-rich tasks where reactivity to small perturbations is key; 
(\textbf{3}) A multi-resolution tactile dataset combining egocentric RGB, vision-based, and event-based tactile sensing across five challenging contact-rich manipulation tasks.

 
    


    
    

    

\section{Related Work}
 \vspace{-0.5em}

\textbf{Tactile Sensing for Robotic Manipulation.}
Touch feedback has become increasingly important for contact-rich robotic manipulation~\citep{li2020review}, particularly in settings where local contact state cannot be reliably inferred from vision alone~\citep{dong2021tactile, calandra2018more,  lach2023placetouch, qi2023general, huang20253d, li_see_2022, ablett2024multimodal, helmut_tactile-conditioned_2025, funk_importance_2025}.  
Different families of tactile sensors have been successfully used to improve manipulation policies. Vision-based tactile sensors, such as GelSight \citep{yuan_gelsight_2017}, provide dense spatial information about contact geometry~\citep{chen2025investigatingactivesamplinghardness,boehm2025textureactiv}, while event-based tactile sensors such as Evetac~\citep{funk_evetac_2024} and GelEvent~\citep{gelevent_2025} capture high-frequency contact changes. 


As tactile sensing has gained traction, reactive diffusion policies have incorporated visuotactile feedback for contact-rich manipulation~\citep{xue_reactive_2025}, demonstrating the value of multi-modal conditioning. Large-scale datasets such as Touch100k~\citep{cheng_touch100k_2025} have enabled touch-centric multimodal representation learning by providing diverse vision-language-touch data. Recent work has explored representations that go beyond pixel-level tactile information~\citep{higuera_tactile_2025} and developed tactile-conditioned diffusion policies for force-aware manipulation~\citep{helmut2025feats, helmut_tactile-conditioned_2025}. Pretraining approaches such as VITaL~\citep{george_vital_2024} have demonstrated that visuotactile pretraining benefits both tactile and non-tactile manipulation policies. The principle that "touch begins where vision ends"~\citep{zhao_touch_2025} has motivated generalizable policies for contact-rich manipulation. Self-supervised contrastive pretraining~\citep{dave_multimodal_2024} has further advanced multimodal visuotactile representation learning. Deep reinforcement learning with visuotactile feedback~\citep{hansen_visuotactile-rl_2022} has enabled learning of complex manipulation policies, while safe self-supervised learning~\citep{fu_safe_2023} has made real-world industrial insertion feasible. Tactile sensing also guides object exploration \citep{yi2016tactileobjectexploration, schneider2025tactilemnistbenchmarkingactive, schneider2026apple}. Recent vision-tactile-language-action models~\citep{zhang_vtla_2025} incorporate preference learning for insertion tasks, and world models such as DreamTacVLA~\citep{ye_learning_2025} enable learning to predict future tactile feedback in simulation~\citep{nguyen2024tacexgelsighttactilesimulation} and real world. Unified representations across static and dynamic tactile sensors~\citep{feng_anytouch_2025} have further demonstrated the potential of multisensory tactile learning. Along similar lines, transferable tactile transformers~\citep{zhao_transferable_2024} have shown promise for learning representations across diverse sensors and tasks.
Despite these significant advancements in tactile-driven manipulation, most current frameworks rely on a single touch sensor modality. Drawing inspiration from the multimodal nature of human touch, this work bridges this gap by exploring how fusing two distinct tactile sensors with complementary temporal resolutions can improve the robustness of learning robotic manipulation.
 
\textbf{Multisensory Learning for Manipulation.}
The integration of vision and touch has been explored extensively for improving manipulation capabilities~\citep{qi2023general, ablett2024multimodal, lenz_analysing_2024, lee_making_2018, lee_making_2019}. Fusing multiple sensors with stage guidance has been applied to complex tasks such as pouring and peg insertion~\citep{feng_play_2024}. The combination of vision and tactile via CNN-stems and masked autoencoding improves visualtactile and vision policies~\citep{sferrazza_power_2023}. Sparsh-X~\citep{higuera_tactile_2025} is training a representation on the multimodal Digit360~\citep{lambeta2024digitizingtouchartificialmultimodal} sensor via Self-
Supervised Learning. The Fusion of Vision, Audio and Touch via a self-attention layer improves single-modality performance~\citep{li_see_2022, zhao2025polytouch}. Visual-force-tactile fusion~\citep{jin_visual-force-tactile_2024} has enabled gentle intricate insertion tasks by leveraging complementary information from multiple sensing modalities. Multi-modal human tactile demonstrations~\citep{yu_mimictouch_2024} have been leveraged to improve contact-rich manipulation through imitation learning. Nevertheless, these approaches focus on cross-modal integration (e.g., vision plus audio) rather than on interplay among tactile-sensing modalities. Despite the promise of multisensory learning, fusing co-located tactile modalities with complementary temporal resolutions to condition manipulation policies remains largely unexplored, a gap that MiTaS directly addresses.




\section{Multi-Resolution Tactile Sensing}

In this section, we present Multi-Resolution Tactile Sensing (MiTaS), a novel framework that leverages multiple heterogeneous tactile sensors operating at different temporal resolutions to learn manipulation policies through imitation.
Our proposed approach consists of three main components: (1) a effective architecture utilizing modality-specific CNN stems that encode diverse sensor inputs into unified token representations, (2) a flow-matching policy for action generation and visuomotor control conditioned on these tokens, and (3) a multi-tactile dropout scheme that leverages all available sensors during training to improve downstream policy performance when only a subset of sensors is available at inference time. To the best of our knowledge this is the first work to combine a classic vision-based tactile sensor with an event-based tactile sensor, showing how complementary tactile features enable complex contact-rich manipulation. Figure~\ref{fig:overview} provides an overview of the MiTaS encoder architecture and policy training.

For training the multisensory visuomotor policies we consider three complementary sensing modalities: egocentric RGB from a wrist-mounted camera, GelSight Mini tactile images~\cite{yuan_gelsight_2017}, and high-rate Evetac tactile frames~\cite{funk_evetac_2024}. At each timestep, the observation is $o_t = \bigl[ O_t^{\text{vision}},\, O_t^{\text{gelsight}},\, O_t^{\text{evetac}} \bigr]$,
where \(O_t^{\text{vision}}~\in~\mathbb{R}^{2 \times 3 \times 128 \times 128}\) is recorded at 25 Hz and captures the scene and hand-object configuration, \(O_t^{\text{gelsight}}~\in~\mathbb{R}^{2 \times 3 \times 120 \times 160}\) is recorded at 25 Hz and captures local contact geometry, and \(O_t^{\text{evetac}}~\in~\mathbb{R}^{16 \times 120 \times 160}\) stacks 16 event-based frames per timestep from the 200 Hz sensor to capture fast contact dynamics.
Vision and GelSight observations are normalized to $[0,1]$, while Evetac is normalized to $[-\tfrac{1}{2},\tfrac{1}{2}]$ so the default grey frame corresponds to $0$ (no event activity) rather than $0.5$.
These input observations are then passed into our proposed MiTaS Architecture, and we subsequently leverage a flow-matching generative model~\cite{black2026pi0, lipman2023flowmatching} for predicting a chunk of end-effector delta actions. We want to highlight that since the policy is neither conditioned on robot state nor expressed in absolute position space, it must infer the appropriate relative motion directly from high-dimensional multimodal sensory input. In the following, we first detail the MiTaS encoding architecture, then the policy parameterization for action generation, and finish with our proposed multi-tactile dropout procedure.

\vspace{-0.25em}
\subsection{Visual Multi-Tactile Encoding Architecture}
\label{subsec:architecture}
\begin{figure}[t]
    \centering
     \includegraphics[width=1.0\textwidth]{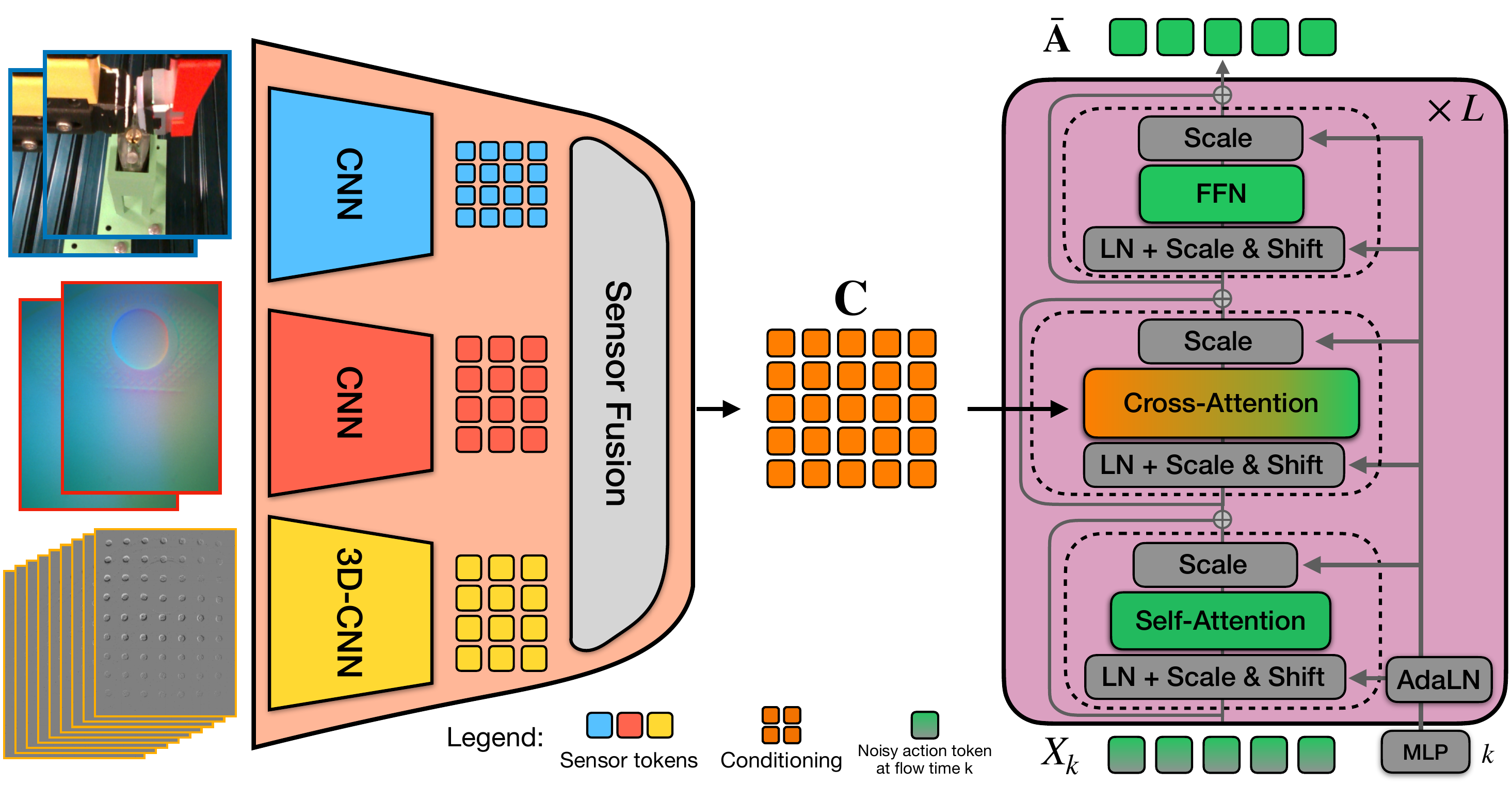}
         \vspace{-2.0em}
        \caption{Overview of the MiTaS architecture: Modality-specific CNN-stems encode Vision, GelSight, and Evetac sensors into token embeddings. These tokens are fused via transformer-based self-attention to condition a flow matching policy. The policy does not receive the robot state and must infer the delta position prediction solely from sensor readings.}
    \label{fig:overview}
    \vspace{-1.5em}
\end{figure}

\vspace{-0.25em}
Extracting and fusing features from heterogeneous sensor modalities is essential to effectively learn multimodal imitation learning policies. We propose a transformer-based architecture to handle multimodal data during encoding and using cross-attention to condition the policy. Cross-attention allows for efficient integration of high-dimensional data, without introducing a bottleneck on the perception side. The Encoder consists of sensor stems, positional / modality encoding, and sensor fusion. An overview of the Architecture is shown in Figure~\ref{fig:overview}. 


\textbf{Sensor Stems:} Each modality is tokenised by a dedicated CNN-stem to extract a meaningful sensor-specific representation. The wrist-camera stacks two RGB frames along the channel axis and applies four strided 2D convolutions, producing a $N_{\mathrm{vis}}=16\times16$ grid of $D{=}256$-dimensional tokens. The GelSight stem mirrors this design at the rectangular $120\times160$ resolution and yields a $N_{\mathrm{gs}}=12\times16$ token grid. Evetac's 16-frame stack is processed by a four-layer 3D-convolutional stem that jointly compresses time and space into a $N_{\mathrm{evt}}=12\times16$ grid, reflecting the higher temporal density of that modality. The 3D-CNN uses temporal kernels of 3–3–3–2 and strides of 2 in time and space through the first three stages, then a final (2$\times$4$\times$5) convolution layer that collapses time to form the token features. The embeddings of the tactile modalities follow the spatial dimension of the sensor observation and allow associations between the different tactile readings later on. Furthermore, early convolutions result in overlapping receptive fields, providing spatially coherent features and have been shown to stabilize training~\cite{xiao2021earlyconvolutionshelptransformers}.

\textbf{Positional and modality encoding:} Each sensor stem outputs a grid of $D$-dimensional tokens per sensor; after flattening, every token $\bm{z}_i$ is augmented with a learned position vector $\bm{p}_{g(i)}$ that encodes its location within that sensor's token grid, and a modality-specific embedding $\bm{e}_{s(i)}$ where $s(i) \in \{\text{vision}, \text{gelsight}, \text{evetac}\}$ identifies the source sensor, i.e., $\bm{z}'_i = \bm{z}_i + \bm{p}_{g(i)} + \bm{e}_{s(i)}$.
This separates \emph{where} a token lies in the spatial layout from \emph{which} modality produced it. Per-token positional encodings and modality-specific segment embeddings are learned jointly with the policy.


\vspace{1.0em}
\textbf{Sensor Fusion:} Rather than pooling each modality into a single global embedding, we retain the full grid of spatially resolved tokens per sensor. The augmented tokens $\bm{z}'_i$ from Vision ($N_{\mathrm{vis}}$), GelSight ($N_{\mathrm{gs}}$), and Evetac ($N_{\mathrm{evt}}$) are concatenated into a visual multi-tactile sequence of $N_{cond}=644$ tokens per observation step, preserving fine-grained visual and tactile detail such as local contact geometry. This sequence is processed by a transformer encoder~\cite{vaswani2023attentionneed} consisting of multi-head self-attention, MLP layers, pre-normalization, and residual connections. The fusion transformer applies full self- and cross-attention across sensors, leading to rich multisensory features that form the conditioning for the policy-head.

\vspace{-1.5em}
\subsection{Action Generation through Flow Matching}
\label{subsec:policytraining}
Next, we utilize the visual multi-tactile embeddings for robot action generation.
Following recent advances in learning robotic manipulation policies from expert demonstrations~\cite{urain2026survey, chi2025diffusion}, we leverage a conditional flow matching generative model for policy learning~\cite{black2026pi0}.
We define our policy output as a fixed-length sequence of future robot commands $\bm{A_t} = [\bm{a}_t, \bm{a}_{t+1}, \ldots, \bm{a}_{t+H-1}] \in\mathbb{R}^{H\times d_a}$, with horizon $H$ and action dimension $d_a$.
The individual actions are \emph{delta} commands in the robot local frame, i.e., $\bm{a} = \Delta\bm{p}=[\Delta x,\Delta y,\Delta z,\Delta\psi]\in\mathbb{R}^{4}$. This formulation controls translation and the yaw rotation around the z-axis, while keeping pitch and roll fixed.
The control loop executes only the first $n_{\mathrm{act}}\le H$ predicted steps before replanning, improving reactivity while retaining a temporally coherent training target.
The policy head is a Diffusion-Transformer (DiT) that regresses a time-conditioned velocity field along a straight noise-to-action path.

\vspace{1.0em}
\textbf{Flow-matching training and inference:}
During training, we start with sampling an initial random noise vector $\bm{x}_0\sim\mathcal{N}(\bm{0},\bm{I})$, and we also sample a desired normalized action chunk from our available demonstrations, i.e., $\bm{x}_1=\bar{\bm{A}}$. Following the conditional flow-matching paradigm~\citep{lipman2023flowmatching}, we define a straight-line probability path $\bm{x}_k$ and its corresponding constant target velocity field $\bm{u}_k$ for $k \in [0,1]$ as: $\bm{x}_{k}=k \cdot \bm{x}_{1}+(1{-}k) \cdot\bm{x}_{0}$, and $\bm{u}_{k}=\mathrm{d}\bm{x}_{k}/\mathrm{d}t$.
The training objective is to optimize the network parameters $\theta$ such that its output $\hat{\bm{v}}_\theta(\bm{x}_k,{k},\bm{C})$ that is conditioned on the current intermediate action state $\bm{x}_k$, the flow time $k$, and the vision and multi-tactile tokens $\bm{C}$ accurately regresses this target velocity field, i.e., we minimize mean-squared error
$
\,\|\hat{\bm{v}}_{\theta}(\bm{x}_k,{k},\bm{C})-\bm{u}_{k}\|^2$ over all the flow time steps across all demonstrations.
During inference, we generate actions by integrating the learned velocity field via explicit Euler integration starting from an initial noise vector, $\bm{x}_0\sim\mathcal{N}(\bm{0},\bm{I})$ and iterate
$\bm{x}_{k+\Delta k}\leftarrow \bm{x}_k+\Delta {k}\,\hat{\bm{v}}_\theta(\bm{x}_k,{k},\bm{C})$
for $k \in [0,1)$ with a fixed $\Delta {k}=1/K$.

\vspace{1.0em}
\textbf{DiT action head:}
The denoiser network ultimately outputs $\hat{\bm{v}}_\theta(\bm{x}_k,{k},\bm{C})$, thereby operating on an intermediate action state $\bm{x}_{k}\in\mathbb{R}^{H\times d_a}$. Each horizon slot is linearly projected to a $D$-dimensional token and summed with a learned positional embedding over the chunk, yielding $\bm{X}_{k}\in\mathbb{R}^{H\times D}$. 
A stack of $L$ DiT blocks updates $\bm{X}_{k}$ while conditioning on multisensory tokens $\bm{C}\in\mathbb{R}^{N_{\mathrm{cond}}\times D}$ produced by the perception encoder; queries come from the action tokens and keys/values from $\bm{C}$, and we use cross-attention without an intermediate projection since both streams share dimension $D$.
Each block applies (i) multi-head self-attention over the $H$ action tokens, (ii) multi-head cross-attention to $\bm{C}$, and (iii) a token-wise MLP, using residual connection throughout. We condition each block on $t$ via AdaLN-Zero~\citep{peebles2023scalablediffusionmodelstransformers, dasari2024ingredientsroboticdiffusiontransformers} to provide a strong, stable time-dependent modulation while initializing the network near an identity mapping (details in Appendix~\ref{app:adalnzero}). The last layer maps the updated tokens back to the final output $\hat{\bm{v}}_\theta(\bm{x}_{k},k,\bm{C})\in\mathbb{R}^{H\times d_a}$.

\subsection{Multi-Tactile Dropout}
Recent works have shown that training a policy on multisensory data can improve vision-based policies, either with pretraining~\citep{sferrazza_power_2023, george_vital_2024, krohn_msdp26} or masking observations
~\citep{funk_importance_2025}. Building on these insights, we introduce a Multi-Tactile Dropout (MTD) scheme designed to train a visuotactile policy (utilizing vision and GelSight) while exploiting the high-frequency Evetac sensor exclusively during the training phase. Each training batch is constructed by sampling half of the sequences with Evetac observations and the remaining half without. This dropout procedure allows the high-frequency temporal dynamics captured by Evetac to shape the representation space of the sensor-fusion network. Crucially, because the model must simultaneously learn to generate accurate actions in the absence of Evetac, this procedure effectively regularizes the standard visuotactile features to compensate for the missing features extracted from Evetac observations implicitly. The Evetac encoder is omitted entirely during deployment, leaving a streamlined vision-and-gelsight policy that benefits from cross-modal feature alignment established during the training phase.

\section{Experimental Evaluation}
 
We evaluate MiTaS on five challenging contact-rich manipulation tasks that require tactile feedback: Gear assembly, Board Wiping, Lamp Installation, Key in Lock, and Lightbulb Connection. Our experiments investigate: (1) overall performance compared to a strong multisensory baseline and vision-only approaches, (2) the advantage of our fusion architecture over two alternative fusion strategies, (3) the contribution of individual sensors through ablation studies, (4) the impact of Multi-Tactile Dropout, and (4) the dynamic importance of different sensors throughout task execution. 

\textbf{Hardware Platform and Data Collection:} This Project uses a 7-DoF Franka Panda robot with an attached Robotis RH-P12-RN gripper. All sensors are mounted on the gripper with 3D-printed holders. Vision is obtained by a wrist-mounted Intel RealSense D435 camera. Gelsight Mini and Evetac are attached to one finger each and form the parallel gripper. Our pipeline is built on Nautilus~\cite{jin2026nautilus} and controls the robot with a Cartesian impedance controller. We collect 30 expert demonstrations by teleoperating the robot via a Spacemouse. The control frequency is 15Hz, and sensor frames are uniformly sampled in a 0.06 s window to form the sensor history per observation step.

\textbf{Baselines and Sensor ablation:} We compare the MiTaS architecture against Sparsh-X~\citep{higuera_tactile_2025}, a multimodal tactile transformer baseline. Sparsh-X uses 8 unimodal and 4 fusion transformer layers, in which attention bottlenecks~\citep{nagrani2022attentionbottlenecksmultimodalfusion} are used as a fusion mechanism instead of self-attention. Instead of pretrained models, we deploy our CNN-stems for PolyTouch~\cite{zhao2025polytouch}, which combines sensor features via cross-attention and uses the class tokens to condition the policy head. Q-former~\citep{li2023blip} fuses CNN-stem sensor features via 32 learnable query-tokens, which then form the multisensory latent space. All models are trained end-to-end with the policy objective using the same DiT action-head. Therefore, the models only differ in architecture and sensor settings to shape the multisensory conditioning of the flow matching policy. For the sensor ablation, we simply omit the encoding of the respective sensor(s). Vision-only models resemble the Vision Transformer architecture~\cite{dosovitskiy2021imageworth16x16words}, following the sensor stems from MiTaS and Sparsh-X, we call them ViT-CNN and ViT.

\subsection{Task Setups}

\begin{figure}[t]
    \centering
    \setlength{\tabcolsep}{3pt}
    
    \begin{tabular}{ccccc}
    \includegraphics[width=0.18\textwidth]{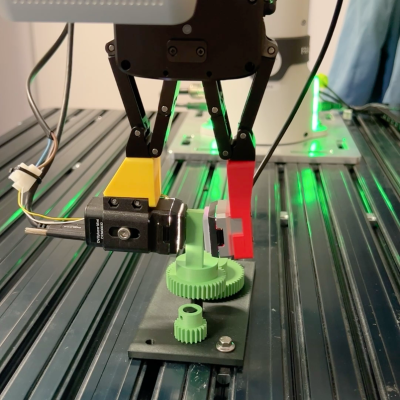} &
    \includegraphics[width=0.18\textwidth]{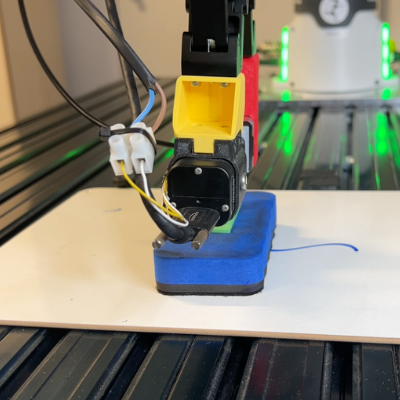} &
    \includegraphics[width=0.18\textwidth]{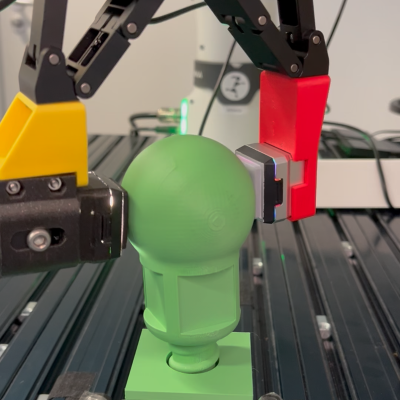} &
    \includegraphics[width=0.18\textwidth]{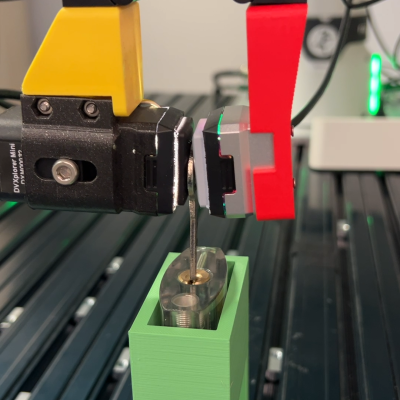} &
    \includegraphics[width=0.18\textwidth]{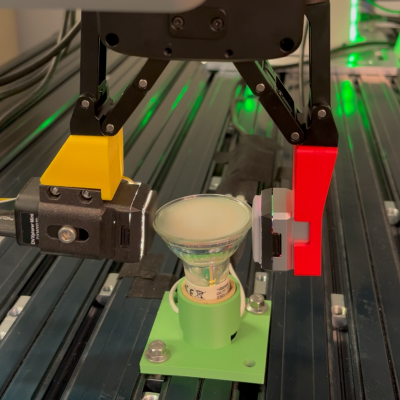} \\
    \end{tabular}
    \vspace{-0.5em}
    \caption{Representative images of the five contact-rich manipulation tasks in order from left to right: Gear Assembly, Board Wiping, Lamp Installation, Key in Lock, and Lightbulb Connection.}
    \label{fig:task_images}
    \vspace{-1.0em}
\end{figure}

\textbf{Gear Assembly}~\cite{tang2023industrealtransferringcontactrichassembly} requires inserting a cylindrical gear between two placed gears. The gear must be aligned in position and orientation to seat on the holder. The gear-teeth can cause it to jam, which is resolved by adjusting orientation and applying fine-grained position changes. \textbf{Board Wiping} entails wiping a marked line from a fixed board. Contact must be maintained along the line throughout the wipe. Insufficient pressure fails to remove the mark, while excessive pressure tilts or wedges the sponge and breaks the planar contact. \textbf{Lamp Installation}~\cite{heo2023furniturebench} requires to align a 3D-printed bulb in a socket in order to screw its threaded base into it. Once the threads are engaged, the bulb is rotated about its vertical axis to drive the threads. Loss of axial alignment can cause it to jam before seating. \textbf{Key in Lock} involves inserting a key fully into a real lock. The key must align with the keyway in both position and orientation. The long and tight insertion path causes even small angular deviations to bind the key against the internal walls. Precise alignment must hold across the full movement until the key is seated. \textbf{Lightbulb Connection} consists of grasping a GU10 LED bulb and connecting it to its socket. The two posts at the base of the bulb must align with the socket slots to enter, after which the bulb is rotated about its vertical axis until the post heads slide into their locking slots. See Figure~\ref{fig:task_images} for an overview of the tasks and Figure~\ref{fig:task_progress} (Appendix) for full task rollouts.

\subsection{Results}

\begin{figure}[h!]
  \centering
      \vspace{-1.0em}
      \includegraphics[width=\textwidth]{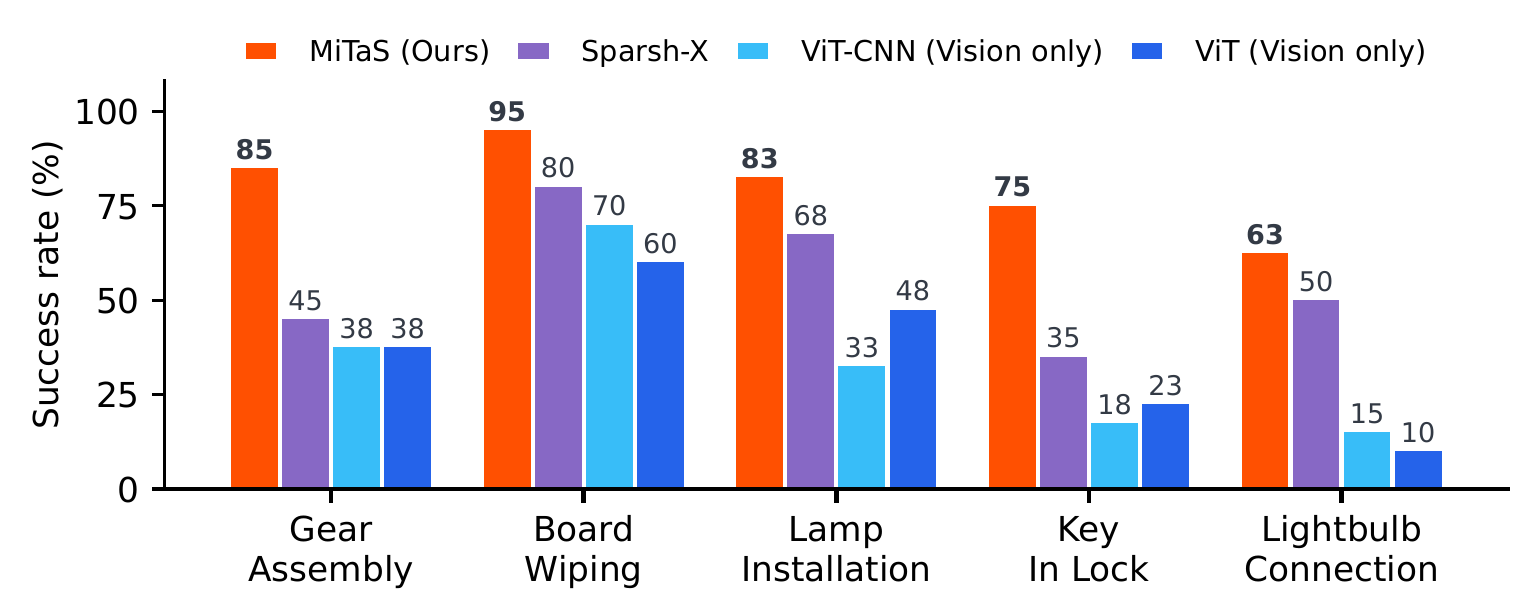}
    \vspace{-2.0em}
  \caption{MiTaS outperformes the multimodal baseline Sparsh-X and two vision-only ViT-baselines in every task. Vision-only models failed, due to occlusion and complex contact dynamics, highlighting the need for tactile sensing in contact-rich manipulation.}

    \vspace{-0.5em}
  \label{fig:main_results}
\end{figure}

We first evaluate overall task performance against the multisensory baseline Sparsh-X and vision-only models. Next, we conduct a sensor fusion study that compares our fusion transformer against alternative strategies for combining visual and tactile features. Followed by a sensor ablation study to showcase the benefits of multi-resolution tactile sensing compared to visuotactile models. To isolate the effect of sensor encoding, all models share the same flow-matching policy head and are trained on the same 30 demonstrations, differing only in which and how the sensor inputs are encoded into conditioning tokens. Lastly, we evaluate our multi-tactile dropout scheme against the sensor ablation from above. We evaluate every policy over 20 trials and report the success rate. The main results in Figure~\ref{fig:main_results} are averaged over two independent datasets, totaling 40 trials per model, which yields a more robust comparison. A table with all results can be found in the Appendix~\ref{subsec:results_overview}.



Figure~\ref{fig:main_results} shows the success rates of MiTaS, Sparsh-X, and both vision-only variants ViT and ViT-CNN over five manipulation tasks. MiTaS, with an average success rate of 80\%, is able to solve all tasks consistently, while Sparsh-X (avg. 55.5\%), with the same sensor setting, is not reliable in every task. Remarkably, Sparsh-X inserted the key multiple times but never completed the task, getting stuck in the keyway. Both vision-only models (ViT 35.5\%; ViT-CNN 34.5\%) have a low performance across all tasks and are not able to consistently solve Key in Lock and Lightbulb Connection, due to tight insertions dynamics and visual occlusions. Multi-resolution tactile feedback paired with suitable encoding enables a high success rate with only 30 demonstrations in these complex contact-rich manipulation tasks.

\begin{wraptable}{r}{0.52\textwidth}
\caption{Sensor Fusion Architecture Study.}
\vspace{-0.5em}
\label{tab:results_fusion}
\centering\small
\renewcommand{\arraystretch}{1.0}
\setlength{\tabcolsep}{4pt}
\begin{tabular}{@{}l*{4}{c}@{}}
\toprule
Task & \textbf{MiTaS} & \textbf{Sparsh-X} & \textbf{Polytouch} & \textbf{Q-former} \\
\midrule
Gear      & \textbf{80\%} & 40\% & 55\% & 50\% \\
Board     & \textbf{100\%} & 75\% & 70\% & 75\% \\
Lamp      & \textbf{85\%} & 65\% & 55\% & 75\% \\
Key       & \textbf{75\%} & 60\% & 55\% & 60\% \\
Lightbulb & \textbf{60\%} & 45\% & 55\% & 55\% \\
\midrule
Avg.      & \textbf{80\%} & 57\% & 58\% & 63\% \\
\bottomrule
\end{tabular}
\vspace{-1.5em}
\end{wraptable}

\textbf{Sensor Fusion Study:} We compare the fusion architecture from our main results against PolyTouch, which uses cross-attention, and Q-Former, which uses learnable queries for sensor fusion. Table~\ref{tab:results_fusion} compares the results of the four sensor fusion architectures across 20 trials, trained on the same dataset using Vision, Gelsight and Evetac as observations. PolyTouch and Q-former have a slightly better average performance than Sparsh-X, while our approach with a high number of conditioning tokens and full self-attention fusion obtains the best performance. The results indicate that, next to the fusion mechanism, the shape of the conditioning is important for imitation learning.

\vfill


\begin{wraptable}[10]{r}[1pt]{0.54\textwidth}
  \centering
  \vspace{-1.0em}
  \footnotesize
  \caption{Sensor Ablation Study over 20 trials.}
  \vspace{-0.5em}
  \label{tab:results_sensor_ablation}
  \renewcommand{\arraystretch}{1.0}  
    \setlength{\tabcolsep}{5pt}
    \footnotesize
    \begin{tabular}{l|ccc|cc}
    \toprule
    & \multicolumn{3}{c|}{\textbf{MiTaS (Ours)}} & \multicolumn{2}{c}{\textbf{Sparsh-X}} \\
    \cmidrule(lr){2-4}\cmidrule(lr){5-6}
    \textbf{Task} & V{+}G{+}E & V{+}G & V{+}E & V{+}G{+}E & V{+}G \\
    \midrule
    Gear        & \textbf{90\%} & 45\% &  0\% & 50\% & 25\% \\
    Board         & \textbf{90\%} & 70\% & 80\% & 85\% & 50\% \\
    Lamp    & \textbf{80\%} & 65\% & 65\% & 70\% & 55\% \\
    Key          & \textbf{75\%} & 55\% &  0\% & 10\% & 20\% \\
    Lightbulb & \textbf{65\%} & 20\% & 40\% & 55\% & 15\% \\
    \midrule
    \textbf{Avg.}        & \textbf{80\%} & 51\% & 37\% & 54\% & 33\% \\
    \bottomrule
    \end{tabular}
      \vspace{-1.5em}
\end{wraptable}


\textbf{Sensor Ablation Study:} We compare the full sensor setting against the Vision and Gelsight setting. Additionally, we have trained the MiTaS model on Vision and Evetac investigating its performance without Gelsight grounding. Table~\ref{tab:results_sensor_ablation} shows the sensor ablation results. Visual multi-tactile models achieve the highest success rate regardless of model design, confirming that heterogenous tactile sensors provide complementary information required to solve contact-rich manipulation task. Still, simple vision-tactile policies are able to solve most of the task over half of the time. Evetac as the sole tactile sensor provides useful information in certain task, but fails completely when object in hand location is needed. Gelsight, on the other hand, can align the gear and key from various initial states (see Appendix~\ref{subsec:gelsight_reset}).

\vfill


\begin{wraptable}{hr}{0.54\textwidth}
 \vspace{-1.0em}
  \centering
  \footnotesize
    \caption{Multi-Tactile Dropout Ablation at V{+}G.}
  \vspace{-0.5em}
  \label{tab:results_cotraining}
        \label{tab:fm_cotraining_ablation}
        \renewcommand{\arraystretch}{1.0}
        \setlength{\tabcolsep}{5pt}
        \footnotesize
        \begin{tabular}{l|ccc|ccc}
        \toprule
        \multirow{2}{*}{\textbf{Task}} &
        \multicolumn{3}{c|}{\textbf{MiTaS (Ours)}} &
        \multicolumn{3}{c}{\textbf{Sparsh-X}} \\
        \cmidrule(lr){2-4}\cmidrule(lr){5-7}
        & V{+}G & MTD & $\Delta$ &
          V{+}G & MTD & $\Delta$ \\
        \midrule
        Gear         & 45\% & \textbf{55\%} & {+}10 & 25\% & 15\% & $-$10 \\
        Board          & 70\% & \textbf{85\%} & {+}15 & 50\% & \textbf{85\%} & {+}35 \\
        Lamp     & 65\% & 60\% & $-$5  & 55\% & \textbf{70\%} & {+}15 \\
        Key  & \textbf{55\%} & 20\% & $-$35 & 20\% &  5\% & $-$15 \\
        Lightbulb & 20\% & \textbf{45\%} & {+}25 & 15\% & 10\% &  $-$5 \\
        \bottomrule
        \end{tabular}
      \vspace{-1.0em}
\end{wraptable}

 
\textbf{Multi-Tactile Dropout:} Table~\ref{tab:results_cotraining} shows the impact of our proposed multi-tactile dropout, leveraging the Evetac sensor during training and not requiring it during policy evaluation. It boosted MiTaS performance in 3/5 tasks, while being helpful in 2/5 tasks for Sparsh-X. In tasks, like Key in Lock, the dropout scheme lowers performance as it disturbs Gelsight's features for in-hand localization. In Board Wiping, both visuotactile models benefit from Evetac information obtained through MTD, aligning with the insights from the sensor ablation (MiTaS V+E) that Evetac features are beneficial in such dynamic tasks. Using Evetac during inference on a multi-tactile dropout trained model increases performance slightly (see Appendix~\ref{subsec:results_overview}).

\subsection{Attention Analysis with Sensor Visualization}
\begin{figure}[ht]
    \centering
    \includegraphics[width=\linewidth]{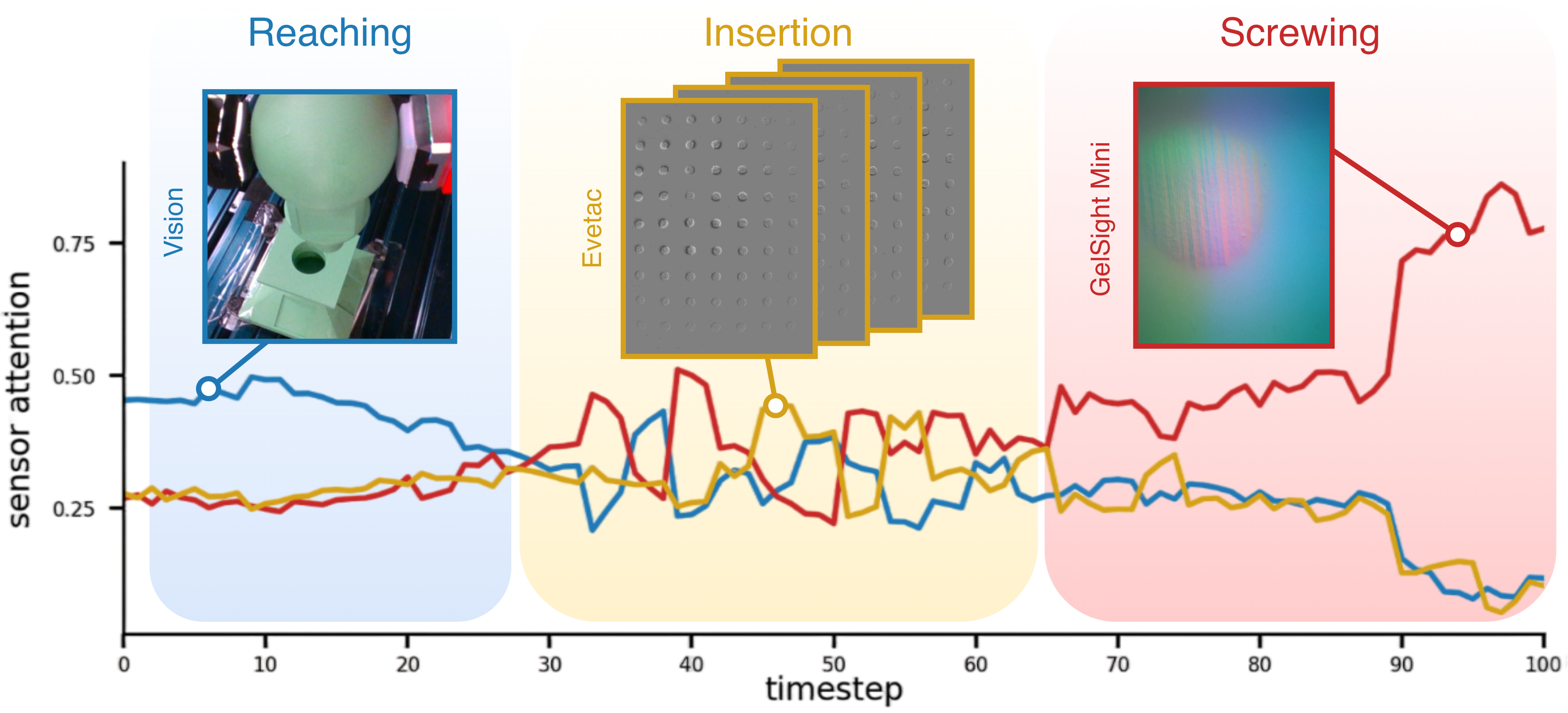}
    \vspace{-1.5em}
    \caption{Attention analysis with sensor readings in Lamp Installation.}
    \label{fig:attention}
    \vspace{-1.0em}
\end{figure}



To understand how MiTaS uses the different sensors over the course of a task, we visualize the cross-attention from the executed action token to each sensor modality during a successful episode of the Lamp Installation task.



Figure~\ref{fig:attention} shows how the policy attends on vision during reaching, on Gelsight for the prolonged screwing motion and on all sensors during insertion. The improved performance and attention show that evetac is not driving the representation, but offers relevant features during crucial stages of the task. We observed recovery behaviour, when it was jammed during Gear assembly and resolving ambiguity in the Lighbulb Installation task, where Gelsight could not differentite between hovering closly over the socket or being fully inserted. Extended Attention analysis can be found in the Appendix~\ref{subsec:attentionanalysis_appen}.


%

\section{Limitations}

While MiTaS demonstrates strong performance across our contact-rich task suite, several limitations remain. 
First, our investigation of multi-resolution fusion focuses exclusively on a specific pair of optical tactile sensors: the frame-based GelSight and the event-based Evetac. While this setup allowed for an exact and detailed analysis of complementary spatial and temporal resolutions, further research is required to evaluate how our architectural framework scales when integrating other heterogeneous sensor pairs or alternative tactile technologies.
Second, our data collection pipeline relies on teleoperation via a Spacemouse, which fundamentally constrains scalability. Transitioning to alternative demonstration sources, such as human video or markerless hand tracking without a physical robot in the loop, could drastically scale up data collection. Furthermore, our policy is currently limited to 4-DoF actions by choice, though the framework itself imposes no fundamental limit on action dimensionality. While a 4-DoF space sufficed for our target contact-rich tasks, scaling to full 6-DoF control would likely require incorporating proprioceptive robot state information, which is currently omitted from our policy's observation space.
Finally, the 15~Hz replanning frequency of our control loop places an upper bound on the framework's reactivity. Integrating our multi-resolution sensing framework with recent advances in highly reactive policy architectures offers a promising direction to further improve policy performance.

\section{Conclusion}
In this work, we propose Multi-Resolution Tactile Sensing (MiTaS), a novel imitation learning framework that leverages multiple tactile sensors operating at different temporal resolutions to solve complex manipulation tasks. Specifically, MiTaS encodes Vision, GelSight, and Evetac frames via CNN-stems and transformer fusion to condition a flow matching policy. Our results demonstrate how vision combined with heterogeneous tactile features obtained at low and high frequencies and spatial resolutions enables fine-grained manipulation, even under heavy occlusion, in complex contact-rich manipulation tasks. MiTaS achieves 80\%, while the next best sensor-fusion architecture has a success rate of 63\% and vision-only policies fail to solve the tasks consistently. This highlights the impact of our multisensory architecture for performing contact-rich tasks. Furthermore, we have shown how Multi-Tactile Dropout can boost a classic visuotactile policy in certain tasks. In future work, we want to investigate event-based tactile features for representation learning on a larger scale to improve widespread vision-based tactile sensors for manipulation. We further plan to investigate more reactive policies that asynchronously fuse vision and high-frequency tactile signals.

\clearpage

\acknowledgments{This research is funded by the German Research Foundation (DFG) Emmy Noether Programme (CH 2676/1-1), the EU’s Horizon Europe project ARISE (Grant no.: 101135959), the German Federal Ministry of Education and Research (BMBF) project “RiG” (Grant no.: 16ME1001) and the European Research Council (ERC) project “SIREN” (Grant No.: 101163933). The authors gratefully acknowledge the scientific support and HPC resources provided by the Erlangen National High Performance Computing Center (NHR@FAU) of the Friedrich-Alexander-Universität Erlangen-Nürnberg (FAU). The hardware is funded by the German Research Foundation (DFG).}


\bibliography{references}  

\newpage
\renewcommand*{\thesection}{\Alph{section}}
\setcounter{section}{0}
\section{Appendix}
This appendix provides supplementary material that supports the experimental evaluation presented in the main text.

\subsection{Controller}
\label{subsec:controller}
We use a Cartesian pose controller implemented as a Cartesian-to-joint target generator coupled with a low-level joint impedance servo. At each control step ($15 \mathrm{Hz}$), the input, provided either by a SpaceMouse teleoperation interface or a learned policy, specifies a target end-effector pose in a reduced Cartesian space $(x, y, z, \psi)$, while roll and pitch are held fixed to values latched at reset. The Cartesian pose error is converted to a bounded Cartesian increment and mapped to joint motion using a Jacobian-based inverse kinematics module with joint-limit handling and nullspace regularization. This yields a desired $7$-DoF joint configuration $\mathbf{q}_d \in \mathbb{R}^7$ that is tracked by a joint impedance controller running at $1 \mathrm{kHz}$, applying $ \tau = K_q(\mathbf{q}_d - \mathbf{q}) + K_{qd}(\dot{\mathbf{q}}_d -\dot{\mathbf{q}})$, resulting in compliant tracking of the commanded Cartesian targets.

\begin{table}[ht]
\centering
\caption{Controller parameters.}
\label{tab:controller_params}
\setlength{\tabcolsep}{8pt}          
\renewcommand{\arraystretch}{1.1}     
\begin{tabular}{@{}l l@{}}
\toprule
Parameter & Value \\
\midrule
Outer-loop command rate & 15~Hz \\
Command space & Cartesian position (xyzyaw) \\
Max linear increment per step & 0.075~m \\
Max rotational increment per step & 0.15~rad \\
Max joint increment per step & 0.2~rad (per joint) \\
Nullspace gain & 0.025 \\
Regularization weight & $1\times 10^{-2}$ \\
Max Cartesian velocity-IK iterations & 300 \\
Max nullspace iterations & 300 \\
Min.\ distance to joint limits & 0.3~rad \\
Impedance servo rate & 1000~Hz \\
Joint stiffness $K_q$ & $[40,\,30,\,50,\,25,\,35,\,25,\,10]$ \\
Joint damping $K_{qd}$ & $[4,\,6,\,5,\,5,\,3,\,2,\,1]$ \\
\bottomrule
\end{tabular}
\end{table}

\newpage
\subsection{Parameter Count and Inference}
\label{subsec:paramcoundinference}
We evaluate all nine policy variants on an NVIDIA GeForce RTX 4080, measuring inference speed over 100 runs per model and reporting the average runtime with 10 flow-matching integration steps (see Table~\ref{tab:param_count_inference}). For each model we also report the number of trainable parameters, split into the perception encoder (Perc.) and the total model size. Every model is using the same policy head with 13,701,508 parameters.  

\begin{table}[h!]
\centering
\caption{Parameter count and inference benchmark.}
\label{tab:param_count_inference}
\begin{tabular}{llcrrccc}
\toprule
Model & Sensors & MTD & \#params Perc. & \#params Total & Tok/obs & ms & Hz \\
\midrule
ViT-CNN & V & -- & 3,079,232 & 16,780,740 & 260 & 35.5 & 227.4 \\
ViT & V & -- & 4,051,328 & 17,752,836 & 260 & 34.4 & 232.2 \\
Sparsh & V+G & -- & 9,398,744 & 23,100,252 & 448 & 39.2 & 203.5 \\
Sparsh & V+G & $\checkmark$ & 15,500,056 & 29,201,564 & 448 & 38.9 & 208.4 \\
Sparsh & V+G+E & -- & 15,500,056 & 29,201,564 & 640 & 54.6 & 153.4 \\
MiTaS & V+E & -- & 8,769,280 & 22,470,788 & 452 & 35.6 & 223.7 \\
MiTaS & V+G & -- & 5,074,776 & 18,776,284 & 452 & 35.3 & 224.1 \\
MiTaS & V+G & $\checkmark$ & 10,764,824 & 24,466,332 & 452 & 36.1 & 230.1 \\
MiTaS & V+G+E & -- & 10,764,824 & 24,466,332 & 644 & 43.5 & 180.5 \\
\bottomrule
\end{tabular}
\end{table}

\subsection{AdaLN-Zero time conditioning.}
\label{app:adalnzero}
Each DiT block is modulated by the time embedding $\boldsymbol{\tau}(t)$ using AdaLN-Zero~\citep{peebles2023scalablediffusionmodelstransformers, dasari2024ingredientsroboticdiffusiontransformers, tancik2020fourierfeaturesletnetworks}.
For a sublayer input $\mathbf{Y}$, we apply adaptive layer normalization
$\mathrm{AdaLN}(\mathbf{Y};\boldsymbol{\beta},\boldsymbol{\gamma})=\mathrm{LN}(\mathbf{Y})\odot(1+\boldsymbol{\gamma})+\boldsymbol{\beta}$,
where $(\boldsymbol{\beta},\boldsymbol{\gamma})$ are produced from $\boldsymbol{\tau}(t)$.
A per-block MLP outputs $(\boldsymbol{\beta},\boldsymbol{\gamma},g)$ triplets for self-attention, cross-attention, and the MLP, and each residual update is gated as
$\mathbf{Y}\leftarrow \mathbf{Y}+g\,f(\mathrm{AdaLN}(\mathbf{Y};\boldsymbol{\beta},\boldsymbol{\gamma}))$.
We initialize the AdaLN projection and the final output head to zero so the network starts near an identity mapping and learns time-conditioning smoothly.

\newpage
\subsection{Symbols and Hyperparameters}
\label{subsec:symbolsnotation}

\begin{table}[ht]
\centering
\caption{Symbols of the MiTaS pipeline. }
\small
\begin{tabular}{ll}
\toprule
Symbol & Meaning \\
\midrule
$H$ & Action horizon \\
$d_a$ & Action dimension \\
$\mathbf{A}\in\mathbb{R}^{H\times d_a}$ & Action chunk  \\
$\bar{\mathbf{A}}\in\mathbb{R}^{H\times d_a}$ & Normalized action chunk \\
$n_{\mathrm{act}}$ & Number of executed actions per replanning step ($n_{\mathrm{act}}\le H$) \\

$O^{\text{vision}} \in \mathbb{R}^{2 \times 3 \times 128 \times 128}$ & Wrist-camera observation \\
$O^{\text{gelsight}} \in \mathbb{R}^{2 \times 3 \times 120 \times 160}$ & GelSight tactile observation \\
$O^{\text{evetac}} \in \mathbb{R}^{16 \times 120 \times 160}$ & EveTac observation  \\

$\mathbf{x}_0\in\mathbb{R}^{H\times d_a}$ & Initial noise sample, $\mathbf{x}_0\sim\mathcal{N}(\mathbf{0},\mathbf{I})$ \\
$\mathbf{x}_1\in\mathbb{R}^{H\times d_a}$ & Data endpoint for flow matching ($\mathbf{x}_1=\bar{\mathbf{a}}$) \\
$k$ & Continuous flow time, $k\in[0,1]$ \\
$\mathbf{x}_k\in\mathbb{R}^{H\times d_a}$ & On-path state returned by the conditional flow matcher \\
$\mathbf{u}_k\in\mathbb{R}^{H\times d_a}$ & Target conditional velocity returned by the flow matcher \\

$\mathbf{C}\in\mathbb{R}^{N_{\mathrm{cond}}\times D}$ & Conditioning tokens from the multisensory encoder \\
$N_{\mathrm{cond}}$ & Number of conditioning tokens \\
$N_{\mathrm{vis}}$  & Number of Vision tokens \\
$N_{\mathrm{gs}}$ & Number of GelSight tokens \\
$N_{\mathrm{evt}}$  & Number of Evetac tokens \\
$D$ & Token embedding width (shared by $\mathbf{C}$ and action tokens) \\
$\mathbf{z}_i\in\mathbb{R}^{D}$ & Sensor token \\
$\mathbf{p}_{g(i)}\in\mathbb{R}^{D}$ & Learned position embedding for token-grid location $g(i)$ \\
$\mathbf{e}_{s(i)}\in\mathbb{R}^{D}$ & Learned modality embedding for source modality $s(i)$ \\
$s(i)\in\{\text{vision},\text{gelsight},\text{evetac}\}$ & Modality identifier for token $i$ \\
$\mathbf{z}'_i\in\mathbb{R}^{D}$ & Augmented token, $\mathbf{z}'_i=\mathbf{z}_i+\mathbf{p}_{g(i)}+\mathbf{e}_{s(i)}$ \\

$\boldsymbol{\tau}(k)\in\mathbb{R}^{d_k}$ & Time embedding of $k$ \\
$v_\theta(\mathbf{x}_k,k,\mathbf{C})$ & DiT network predicting velocity / vector field \\
$\hat{\mathbf{v}}_\theta\in\mathbb{R}^{H\times d_a}$ & Predicted velocity chunk \\
$m_h$ & Validity mask for timestep $h$ (episode padding mask) \\
$\mathcal{L}_{\mathrm{FM}}$ & Flow-matching training loss (masked MSE) \\
$K$ & Number of Euler integration steps at inference \\
$\Delta k$ & Euler step size, $\Delta k = 1/K$ \\
$\Delta\mathbf{p}\in\mathbb{R}^{4}$ & Delta position and delta yaw: $[\Delta x,\Delta y,\Delta z,\Delta\psi]$ \\
\bottomrule
\end{tabular}

\label{tab:policy_symbols}
\end{table}


\begin{table}[ht]
\centering
\caption{Hyperparameters.}
\label{tab:policy_default_hparams}
\small
\setlength{\tabcolsep}{6pt}
\renewcommand{\arraystretch}{1.05}
\begin{tabular}{ll|ll}
\toprule
Hyperparameter & Value & Hyperparameter & Value \\
\midrule
Batch size & 64
& Vision normalization & $[0,1]$ \\
Action horizon $H$ & $16$
& Gelsight normalization & $[0,1]$ \\
Executed steps per replan $n_{\mathrm{act}}$ & $1$--$3$
& Evetac normalization & $[-\tfrac{1}{2},\tfrac{1}{2}]$ \\
Observation window $n_{\mathrm{obs}}$ & $4$
& Embedding width $D$ & $256$ \\
Action normalization & minmax
& DiT depth $L$ & $10$ \\
Action target & $\Delta\mathbf{p}\in\mathbb{R}^{4}$
& Attention heads & $8$ \\
Time embedding dim $d_t$ & $128$
& MLP ratio & $4$ \\
Time embedding & sinusoidal
& Dropout & $0.0$ \\
Flow-matching noise $\sigma$ & $0.0$
& FM inference steps $K$ & $10$ \\
Policy learning rate & $1\times 10^{-4}$
& Encoder learning rate & $1\times 10^{-5}$ \\
Weight decay & $1\times 10^{-6}$
& Adam betas & $(0.95, 0.999)$ \\
Learning-rate schedule & cosine
& Gradient clipping & $1.0$ \\
\bottomrule
\end{tabular}
\end{table}




\clearpage
\newpage 
\subsection{Task Progress}
\begin{figure}[h!]
    \centering
    \setlength{\tabcolsep}{4pt}
    
    \begin{tabular}{c@{\hspace{8pt}}ccccc}
    
    \adjustbox{valign=c}{\rotatebox{90}{\small\textbf{Gear Assembly}}} &
    \adjustbox{valign=c}{\includegraphics[width=0.17\textwidth]{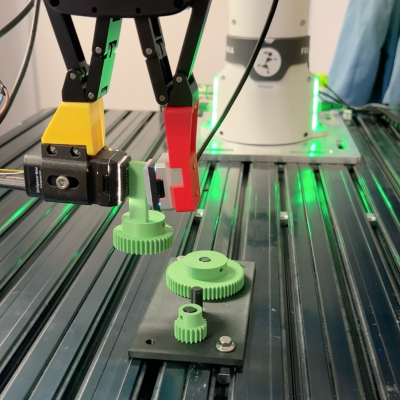}} &
    \adjustbox{valign=c}{\includegraphics[width=0.17\textwidth]{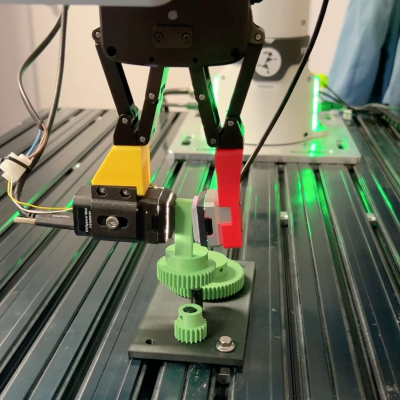}} &
    \adjustbox{valign=c}{\includegraphics[width=0.17\textwidth]{figures/task_progress/gearassembly_3.png}} &
    \adjustbox{valign=c}{\includegraphics[width=0.17\textwidth]{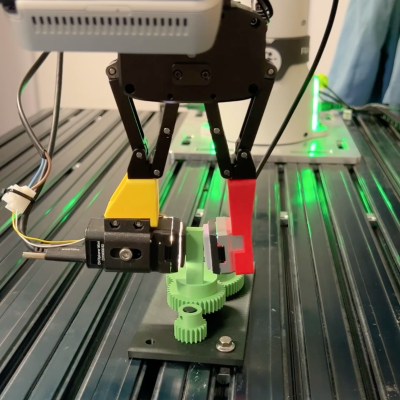}} &
    \adjustbox{valign=c}{\includegraphics[width=0.17\textwidth]{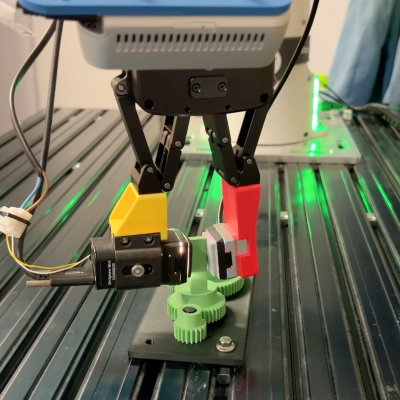}} \\ \addlinespace[4pt]
    & \parbox[t][2.2em][t]{0.17\textwidth}{\centering\small Initial State}
    & \parbox[t][2.2em][t]{0.17\textwidth}{\centering\small First Contact}
    & \parbox[t][2.2em][t]{0.17\textwidth}{\centering\small Alignment}
    & \parbox[t][2.2em][t]{0.17\textwidth}{\centering\small Adjustment}
    & \parbox[t][2.2em][t]{0.17\textwidth}{\centering\small Gear assembled} \\
    \addlinespace[0pt]
    
    \adjustbox{valign=c}{\rotatebox{90}{\small\textbf{Board Wiping}}} &
    \adjustbox{valign=c}{\includegraphics[width=0.17\textwidth]{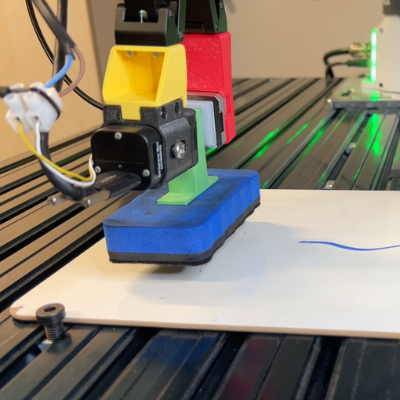}} &
    \adjustbox{valign=c}{\includegraphics[width=0.17\textwidth]{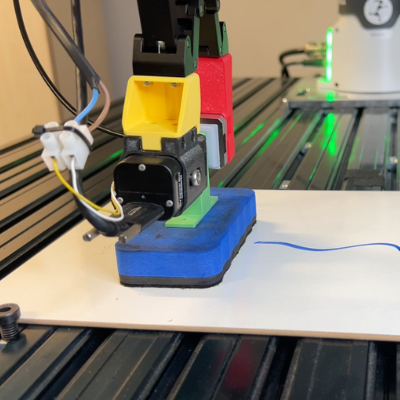}} &
    \adjustbox{valign=c}{\includegraphics[width=0.17\textwidth]{figures/task_progress/boardwiping_3.png}} &
    \adjustbox{valign=c}{\includegraphics[width=0.17\textwidth]{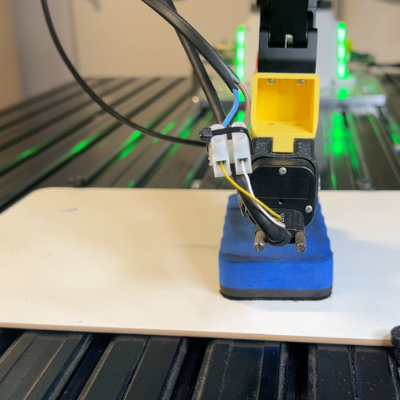}} &
    \adjustbox{valign=c}{\includegraphics[width=0.17\textwidth]{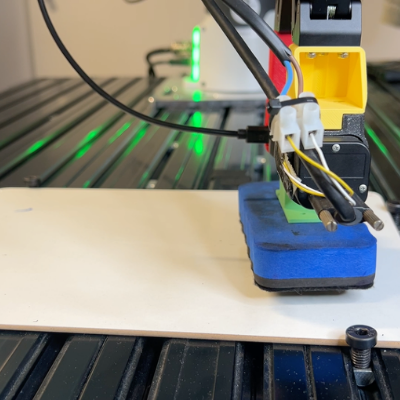}} \\ \addlinespace[4pt]
    & \parbox[t][2.2em][t]{0.17\textwidth}{\centering\small Initial State}
    & \parbox[t][2.2em][t]{0.17\textwidth}{\centering\small First Contact}
    & \parbox[t][2.2em][t]{0.17\textwidth}{\centering\small Wiping}
    & \parbox[t][2.2em][t]{0.17\textwidth}{\centering\small Wiping (cont.)}
    & \parbox[t][2.2em][t]{0.17\textwidth}{\centering\small Board Wiped} \\
    \addlinespace[0pt]
    
    \adjustbox{valign=c}{\rotatebox{90}{\small\textbf{Lamp Installation}}} &
    \adjustbox{valign=c}{\includegraphics[width=0.17\textwidth]{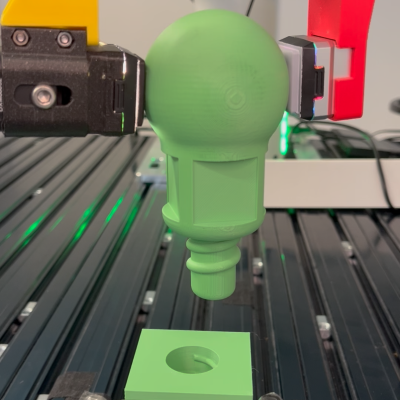}} &
    \adjustbox{valign=c}{\includegraphics[width=0.17\textwidth]{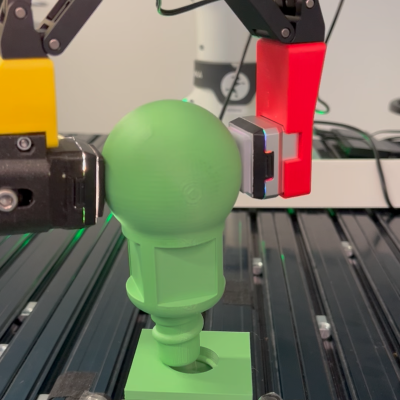}} &
    \adjustbox{valign=c}{\includegraphics[width=0.17\textwidth]{figures/task_progress/lampinstallation_3.png}} &
    \adjustbox{valign=c}{\includegraphics[width=0.17\textwidth]{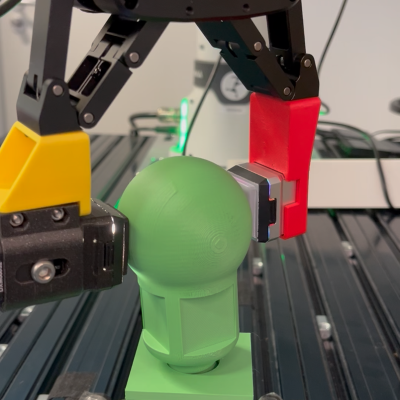}} &
    \adjustbox{valign=c}{\includegraphics[width=0.17\textwidth]{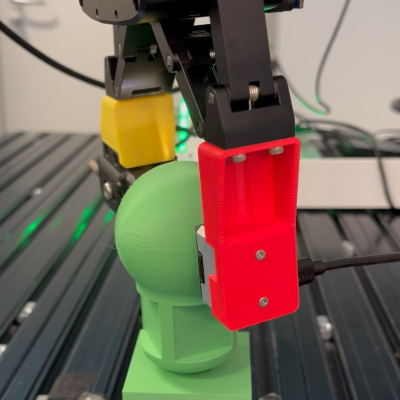}} \\ \addlinespace[4pt]
    & \parbox[t][2.2em][t]{0.17\textwidth}{\centering\small Initial State}
    & \parbox[t][2.2em][t]{0.17\textwidth}{\centering\small First Contact}
    & \parbox[t][2.2em][t]{0.17\textwidth}{\centering\small Alignment}
    & \parbox[t][2.2em][t]{0.17\textwidth}{\centering\small Threading}
    & \parbox[t][2.2em][t]{0.17\textwidth}{\centering\small Lamp Installed} \\
    \addlinespace[0pt]
    
    \adjustbox{valign=c}{\rotatebox{90}{\small\textbf{Key in Lock}}} &
    \adjustbox{valign=c}{\includegraphics[width=0.17\textwidth]{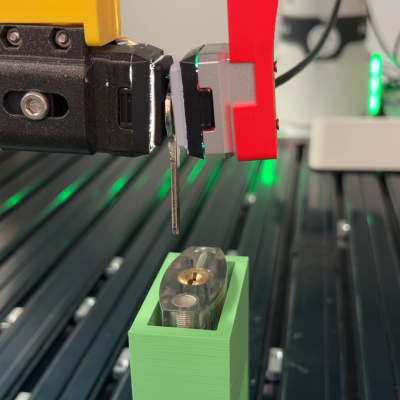}} &
    \adjustbox{valign=c}{\includegraphics[width=0.17\textwidth]{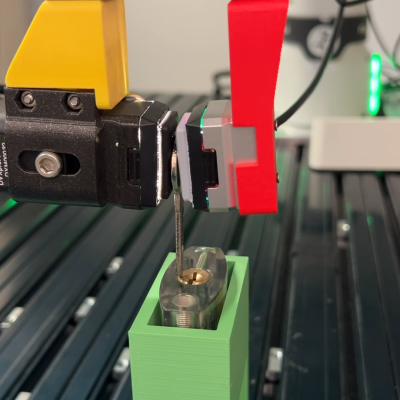}} &
    \adjustbox{valign=c}{\includegraphics[width=0.17\textwidth]{figures/task_progress/keyinlock_3.png}} &
    \adjustbox{valign=c}{\includegraphics[width=0.17\textwidth]{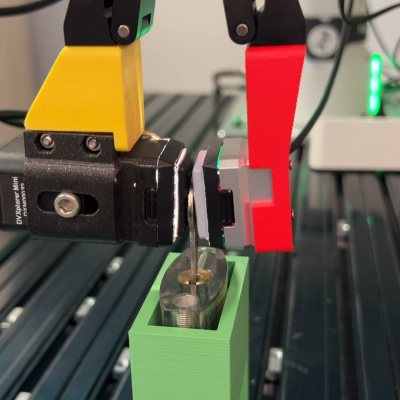}} &
    \adjustbox{valign=c}{\includegraphics[width=0.17\textwidth]{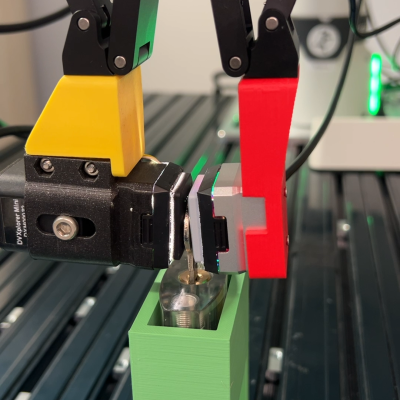}} \\ \addlinespace[4pt]
    & \parbox[t][2.2em][t]{0.17\textwidth}{\centering\small Initial State}
    & \parbox[t][2.2em][t]{0.17\textwidth}{\centering\small First Contact}
    & \parbox[t][2.2em][t]{0.17\textwidth}{\centering\small Alignment}
    & \parbox[t][2.2em][t]{0.17\textwidth}{\centering\small Inserting}
    & \parbox[t][2.2em][t]{0.17\textwidth}{\centering\small Key Inserted} \\
    \addlinespace[0pt]
    
    \adjustbox{valign=c}{\rotatebox{90}{\small\textbf{Lightbulb Connection}}} &
    \adjustbox{valign=c}{\includegraphics[width=0.17\textwidth]{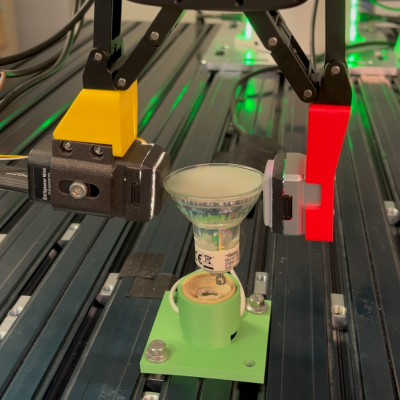}} &
    \adjustbox{valign=c}{\includegraphics[width=0.17\textwidth]{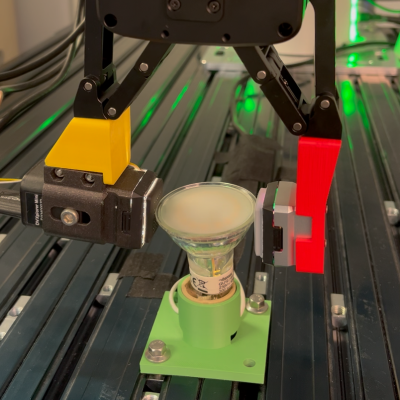}} &
    \adjustbox{valign=c}{\includegraphics[width=0.17\textwidth]{figures/task_progress/lightbulbconnect_3.png}} &
    \adjustbox{valign=c}{\includegraphics[width=0.17\textwidth]{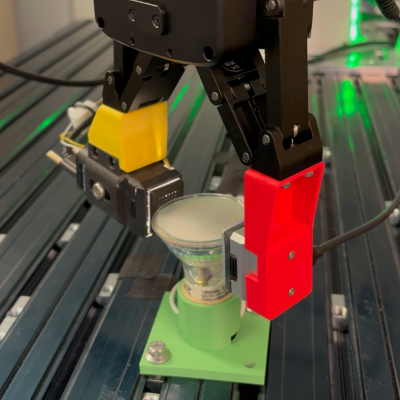}} &
    \adjustbox{valign=c}{\includegraphics[width=0.17\textwidth]{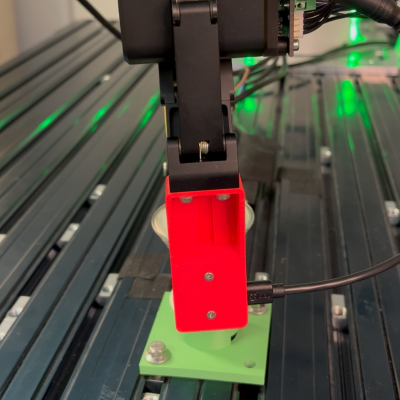}} \\ \addlinespace[4pt]
    & \parbox[t][2.2em][t]{0.17\textwidth}{\centering\small Initial State}
    & \parbox[t][2.2em][t]{0.17\textwidth}{\centering\small First Contact}
    & \parbox[t][2.2em][t]{0.17\textwidth}{\centering\small Alignment}
    & \parbox[t][2.2em][t]{0.17\textwidth}{\centering\small  Rotate in}
    & \parbox[t][2.2em][t]{0.17\textwidth}{\centering\small Bulb Connected} \\
    
    \end{tabular}
    \vspace{-1.0em}
    \caption{Task progress across five manipulation tasks. Each row shows representative frames from a successful rollout, progressing from the initial scene configuration (left) to task completion (right). The tasks span a range of contact-rich skills, including Gear Assembly, Board Wiping, Lamp Installation (FurnitureBench), Key in Lock, and Lightbulb Connection.}
    \label{fig:task_progress}
\end{figure}

\newpage
\subsection{Failure Cases}

\begin{figure}[h!]
    \centering
    \setlength{\tabcolsep}{4pt}
    
    \begin{tabular}{c@{\hspace{8pt}}ccc}
    
    \adjustbox{valign=c}{\rotatebox{90}{\small\textbf{Gear Assembly}}} &
    \adjustbox{valign=c}{\includegraphics[width=0.20\textwidth]{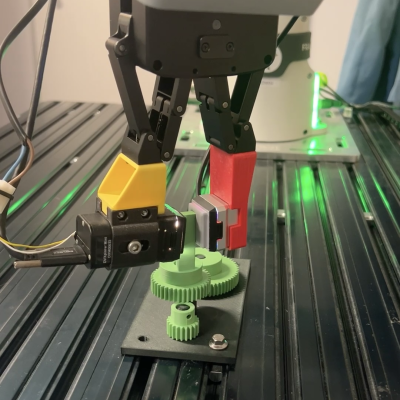}} &
    \adjustbox{valign=c}{\includegraphics[width=0.20\textwidth]{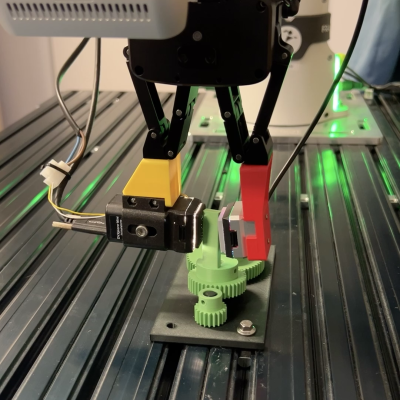}} &
    \adjustbox{valign=c}{\includegraphics[width=0.20\textwidth]{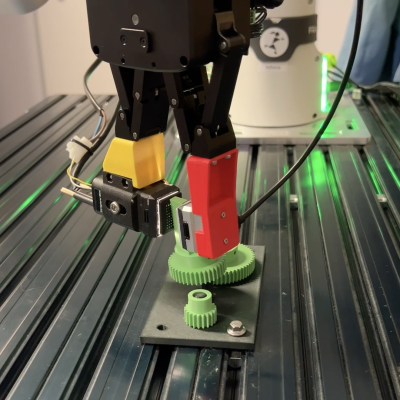}} \\ \addlinespace[4pt]
    & \parbox[t][2.2em][t]{0.20\textwidth}{\centering\small Insertion failure}
    & \parbox[t][2.2em][t]{0.20\textwidth}{\centering\small Tooth jamming}
    & \parbox[t][2.2em][t]{0.20\textwidth}{\centering\small Premature rotation} \\
    \addlinespace[0pt]
    
    \adjustbox{valign=c}{\rotatebox{90}{\small\textbf{Board Wiping}}} &
    \adjustbox{valign=c}{\includegraphics[width=0.20\textwidth]{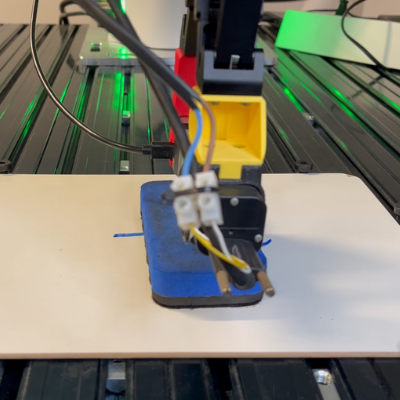}} &
    \adjustbox{valign=c}{\includegraphics[width=0.20\textwidth]{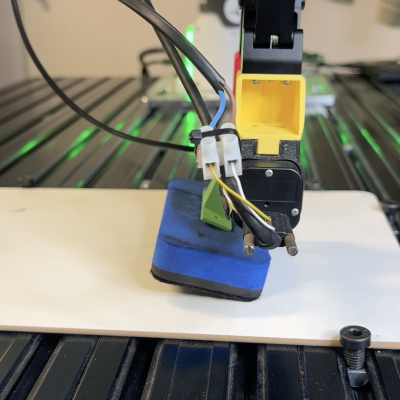}} &
    \adjustbox{valign=c}{\includegraphics[width=0.20\textwidth]{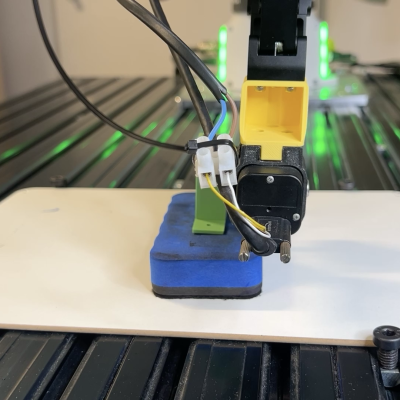}} \\ \addlinespace[4pt]
    & \parbox[t][2.2em][t]{0.20\textwidth}{\centering\small Insufficient pressure}
    & \parbox[t][2.2em][t]{0.20\textwidth}{\centering\small Sponge tilt}
    & \parbox[t][2.2em][t]{0.20\textwidth}{\centering\small Sponge slip-out} \\
    \addlinespace[0pt]
    
    \adjustbox{valign=c}{\rotatebox{90}{\small\textbf{Lamp Installation}}} &
    \adjustbox{valign=c}{\includegraphics[width=0.20\textwidth]{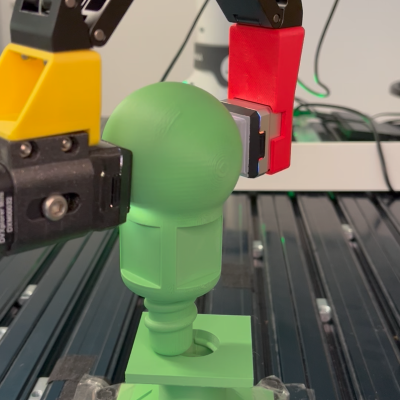}} &
    \adjustbox{valign=c}{\includegraphics[width=0.20\textwidth]{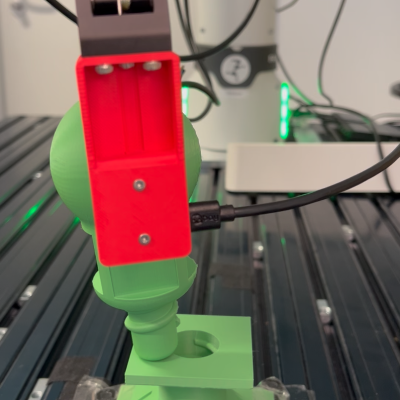}} &
    \adjustbox{valign=c}{\includegraphics[width=0.20\textwidth]{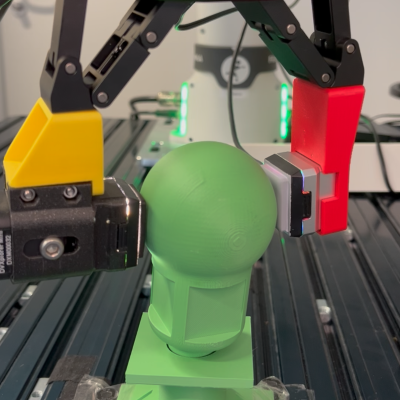}} \\ \addlinespace[4pt]
    & \parbox[t][2.2em][t]{0.20\textwidth}{\centering\small Insertion failure}
    & \parbox[t][2.2em][t]{0.20\textwidth}{\centering\small Premature rotation}
    & \parbox[t][2.2em][t]{0.20\textwidth}{\centering\small Thread jamming} \\
    \addlinespace[0pt]
    
    \adjustbox{valign=c}{\rotatebox{90}{\small\textbf{Key in Lock}}} &
    \adjustbox{valign=c}{\includegraphics[width=0.20\textwidth]{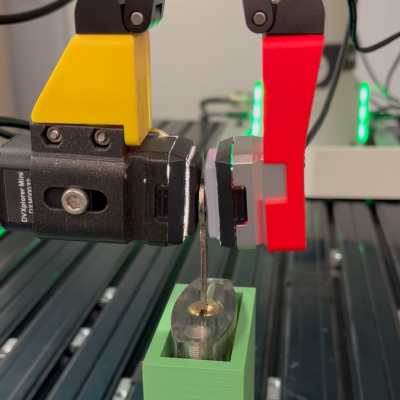}} &
    \adjustbox{valign=c}{\includegraphics[width=0.20\textwidth]{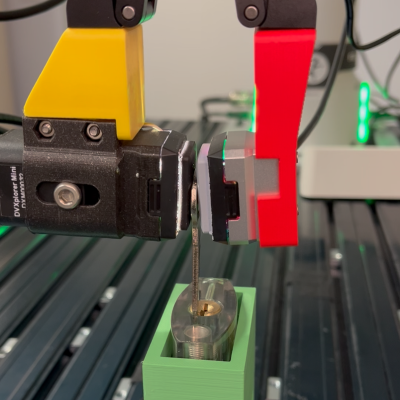}} &
    \adjustbox{valign=c}{\includegraphics[width=0.20\textwidth]{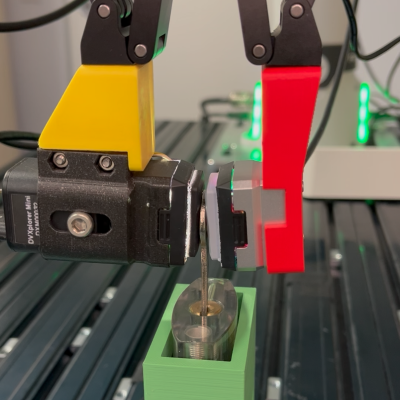}} \\ \addlinespace[4pt]
    & \parbox[t][2.2em][t]{0.20\textwidth}{\centering\small Edge catch}
    & \parbox[t][2.2em][t]{0.20\textwidth}{\centering\small Keyway misalignment}
    & \parbox[t][2.2em][t]{0.20\textwidth}{\centering\small Rim jamming} \\
    \addlinespace[0pt]
    
    \adjustbox{valign=c}{\rotatebox{90}{\small\textbf{Lightbulb Connection}}} &
    \adjustbox{valign=c}{\includegraphics[width=0.20\textwidth]{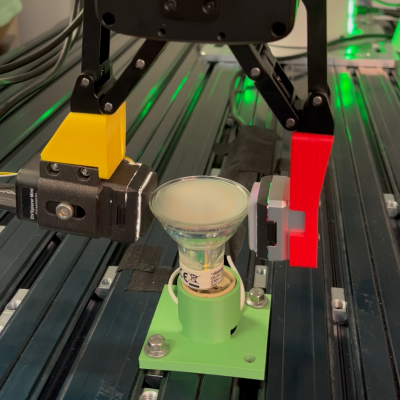}} &
    \adjustbox{valign=c}{\includegraphics[width=0.20\textwidth]{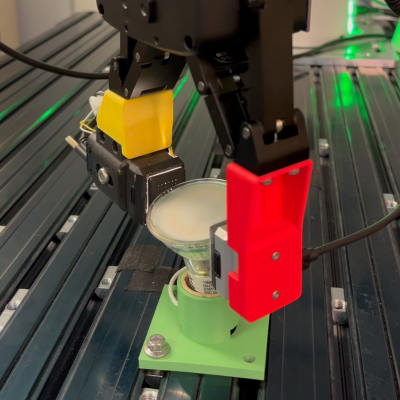}} &
    \adjustbox{valign=c}{\includegraphics[width=0.20\textwidth]{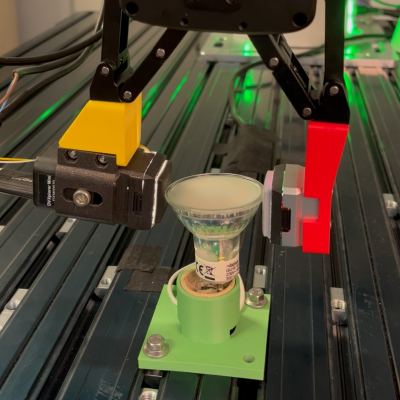}} \\ \addlinespace[4pt]
    & \parbox[t][2.2em][t]{0.20\textwidth}{\centering\small Post misalignment}
    & \parbox[t][2.2em][t]{0.20\textwidth}{\centering\small Single-post insertion}
    & \parbox[t][2.2em][t]{0.20\textwidth}{\centering\small Rim jamming} \\
    
    \end{tabular}
    \vspace{-1.0em}
    \caption{Examples of failure cases across the five manipulation tasks. Each row shows representative frames from a failed rollout. These examples are illustrative and not exhaustive.}
    \label{fig:task_failure}
\end{figure}



\newpage
\subsection{In-Hand Distribution of Initial States}
\label{subsec:gelsight_reset}
\begin{figure}[ht!]
\centering

\newcommand{\montageH}{2.35cm}
\newcommand{\stdH}{2.5cm}
\newcommand{\taskcol}{0.2\textwidth}  

\begin{minipage}[t]{\taskcol}
  \centering
  \begin{subfigure}{\linewidth}
    \centering
    \includegraphics[height=\montageH,width=\linewidth,keepaspectratio]{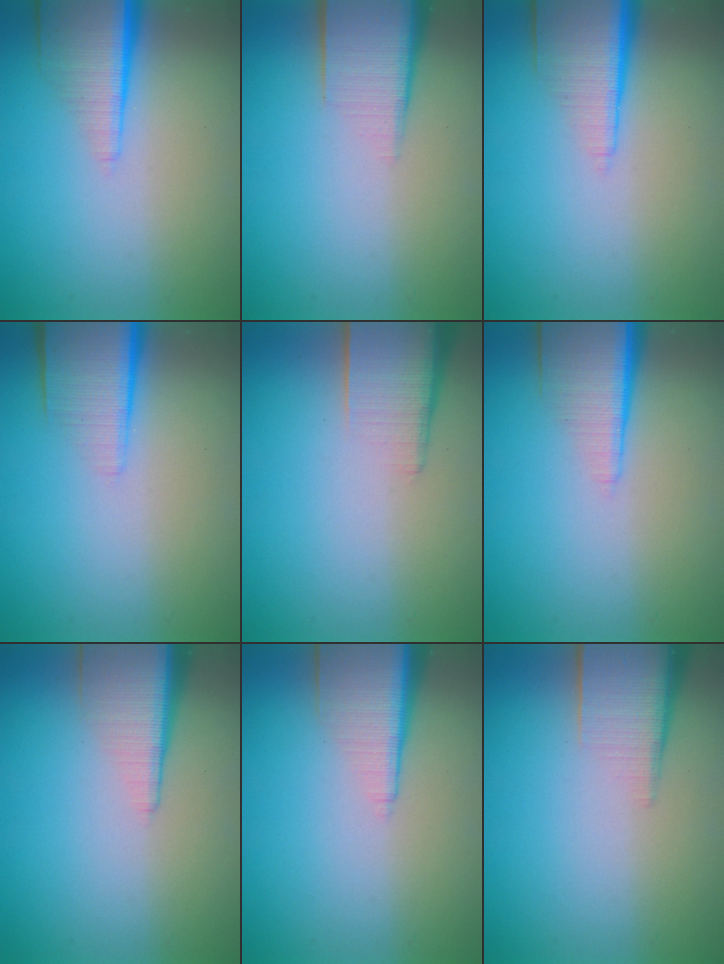}
    \caption{}
  \end{subfigure}
  \begin{subfigure}{\linewidth}
    \centering
    \includegraphics[height=\stdH,width=\linewidth,keepaspectratio]{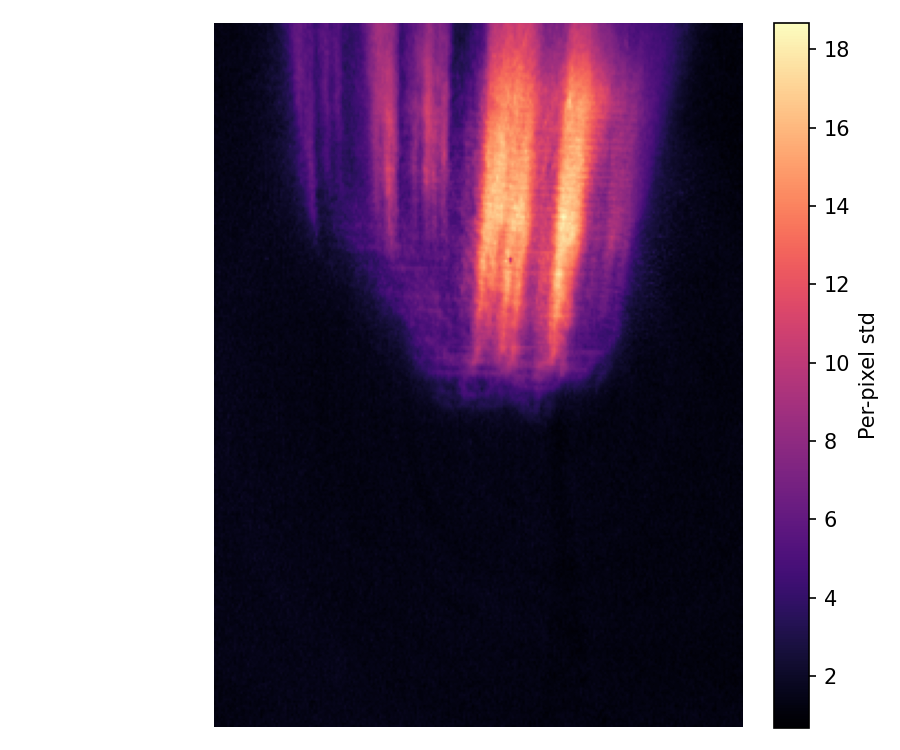}
    \caption{}
  \end{subfigure}
\end{minipage}\hfill
\begin{minipage}[t]{\taskcol}
  \centering
  \begin{subfigure}{\linewidth}
    \centering
    \includegraphics[height=\montageH,width=\linewidth,keepaspectratio]{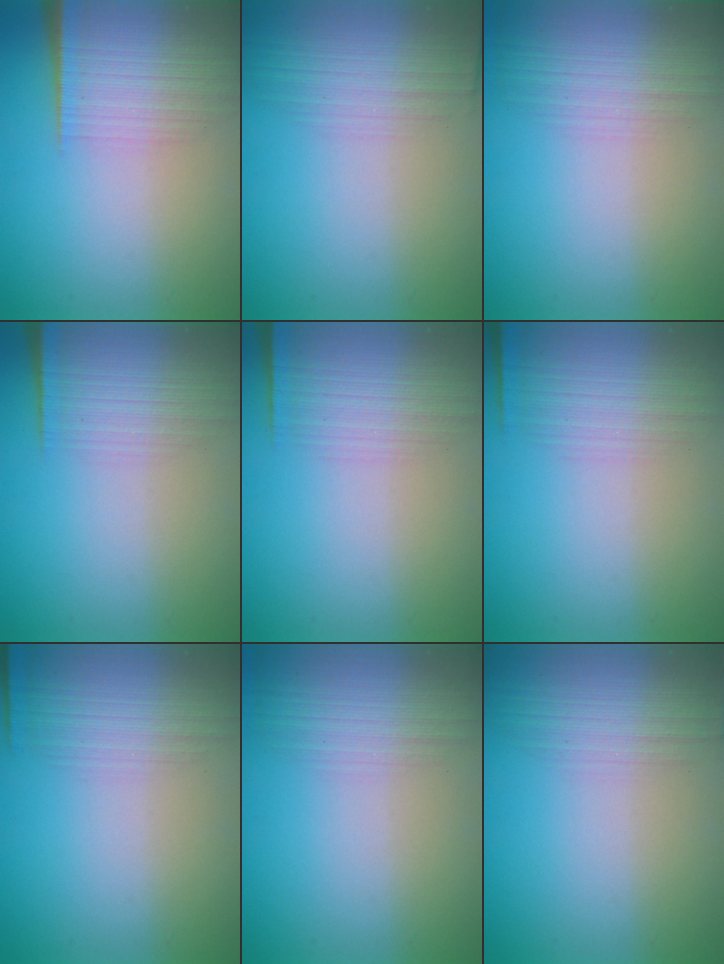}
    \caption{}
  \end{subfigure}
  \begin{subfigure}{\linewidth}
    \centering
    \includegraphics[height=\stdH,width=\linewidth,keepaspectratio]{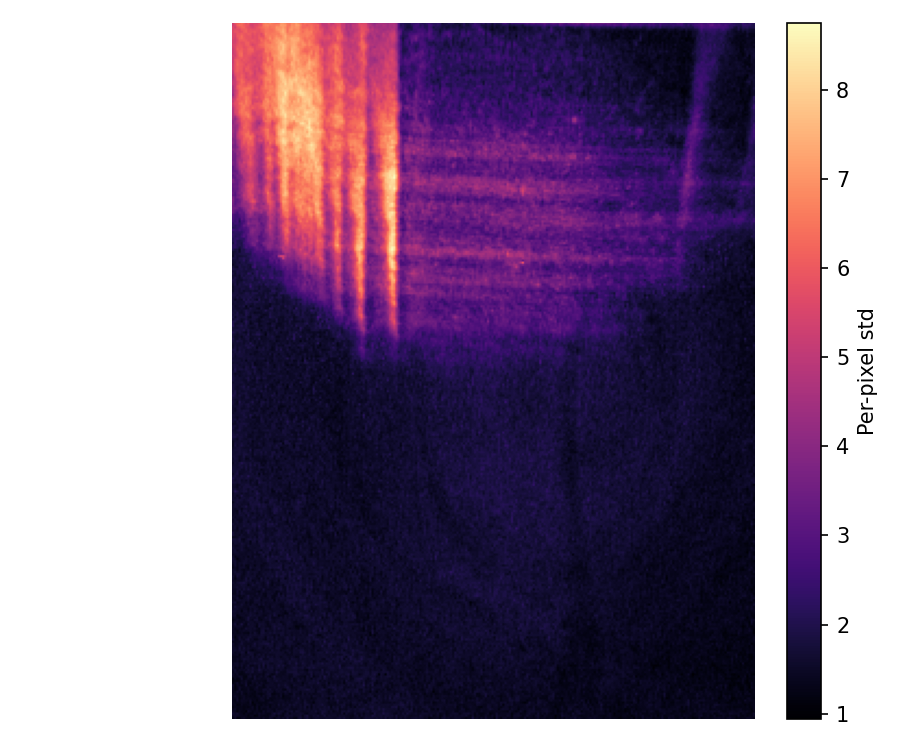}
    \caption{}
  \end{subfigure}
\end{minipage}\hfill
\begin{minipage}[t]{\taskcol}
  \centering
  \begin{subfigure}{\linewidth}
    \centering
    \includegraphics[height=\montageH,width=\linewidth,keepaspectratio]{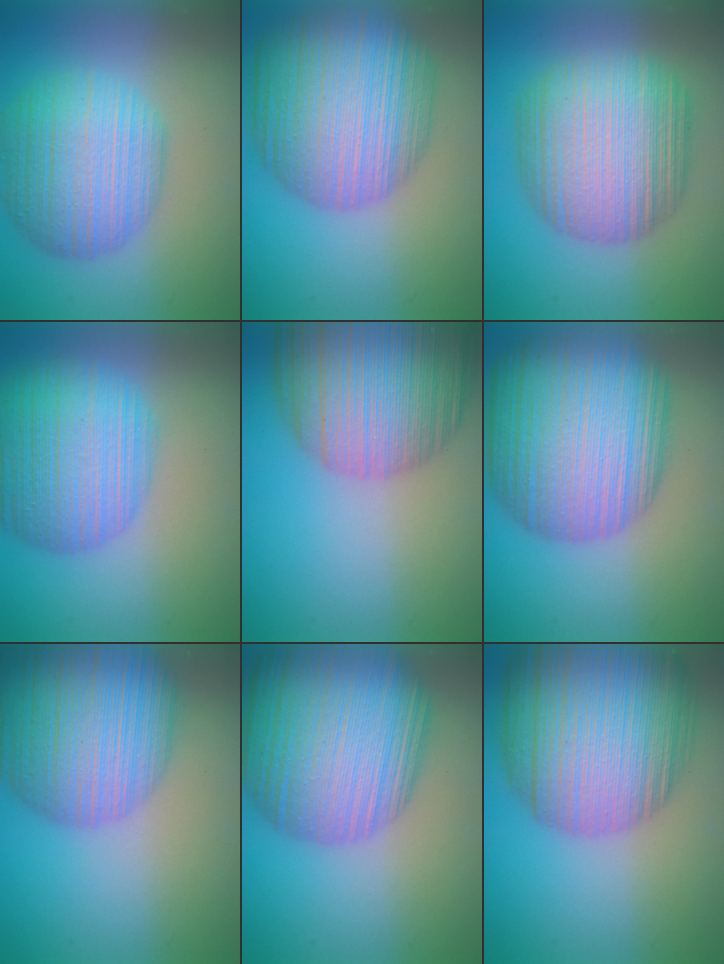}
    \caption{}
  \end{subfigure}
  \begin{subfigure}{\linewidth}
    \centering
    \includegraphics[height=\stdH,width=\linewidth,keepaspectratio]{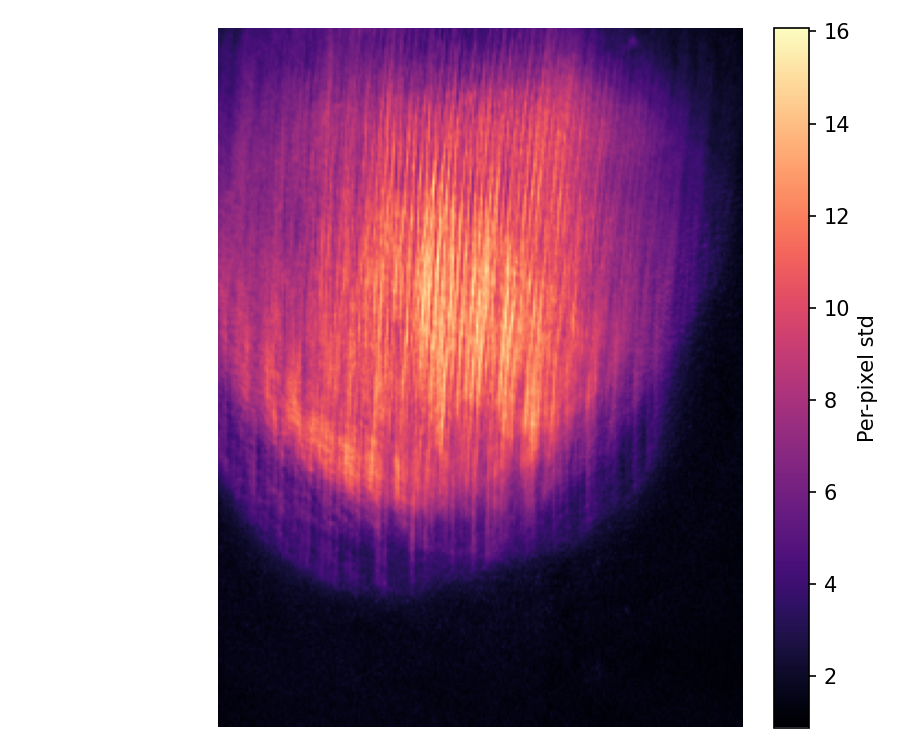}
    \caption{}
  \end{subfigure}
\end{minipage}\hfill
\begin{minipage}[t]{\taskcol}
  \centering
  \begin{subfigure}{\linewidth}
    \centering
    \includegraphics[height=\montageH,width=\linewidth,keepaspectratio]{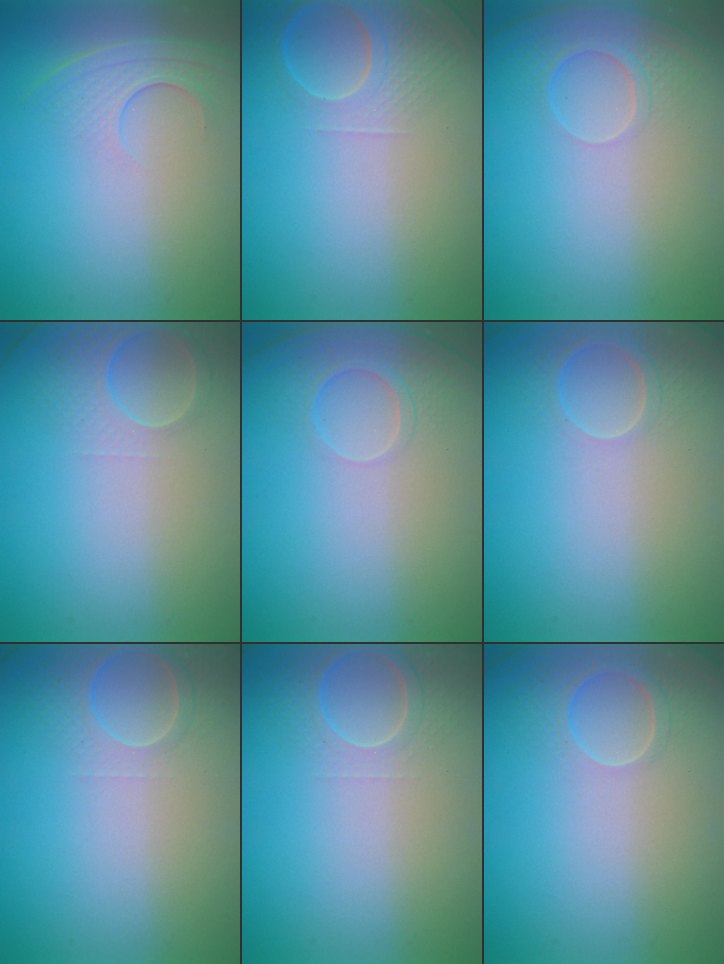}
    \caption{}
  \end{subfigure}
  \begin{subfigure}{\linewidth}
    \centering
    \includegraphics[height=\stdH,width=\linewidth,keepaspectratio]{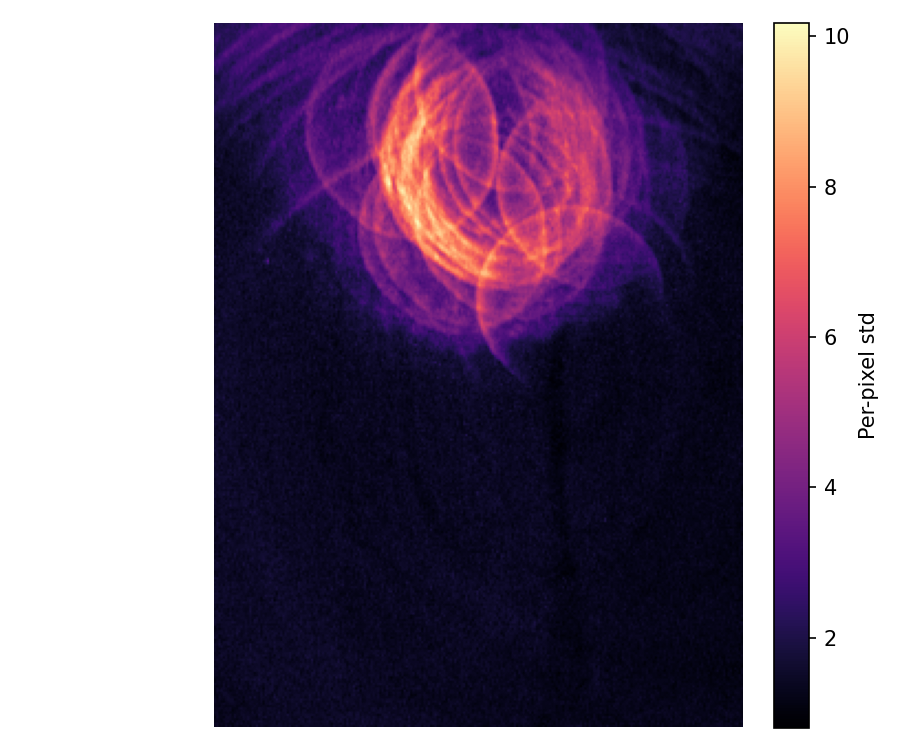}
    \caption{}
  \end{subfigure}
\end{minipage}\hfill
\begin{minipage}[t]{\taskcol}
  \centering
  \begin{subfigure}{\linewidth}
    \centering
    \includegraphics[height=\montageH,width=\linewidth,keepaspectratio]{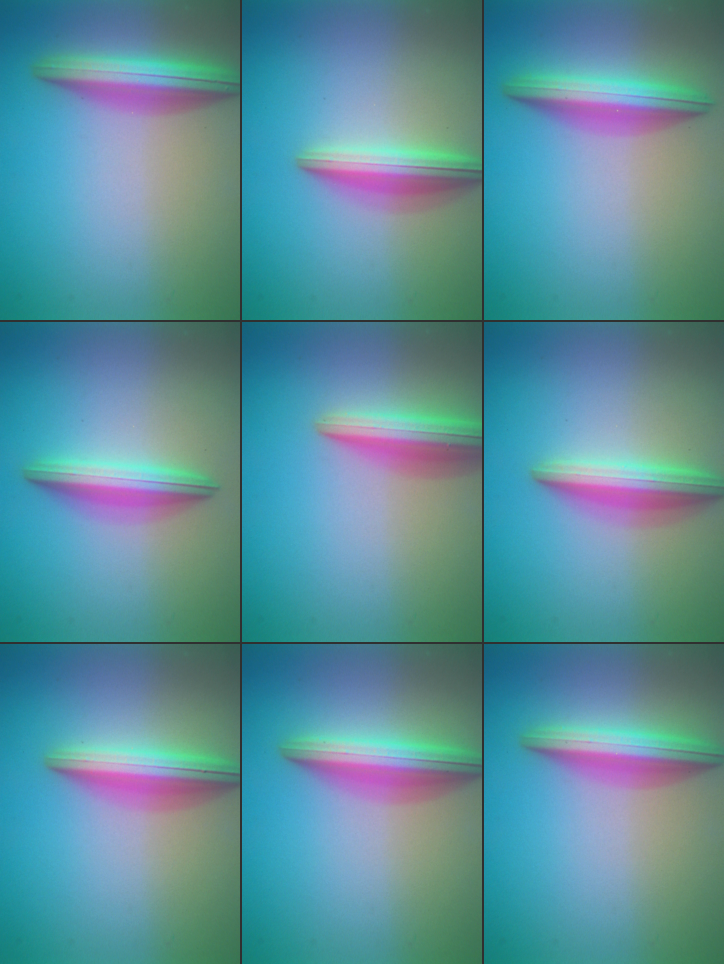}
    \caption{}
  \end{subfigure}
  \begin{subfigure}{\linewidth}
    \centering
    \includegraphics[height=\stdH,width=\linewidth,keepaspectratio]{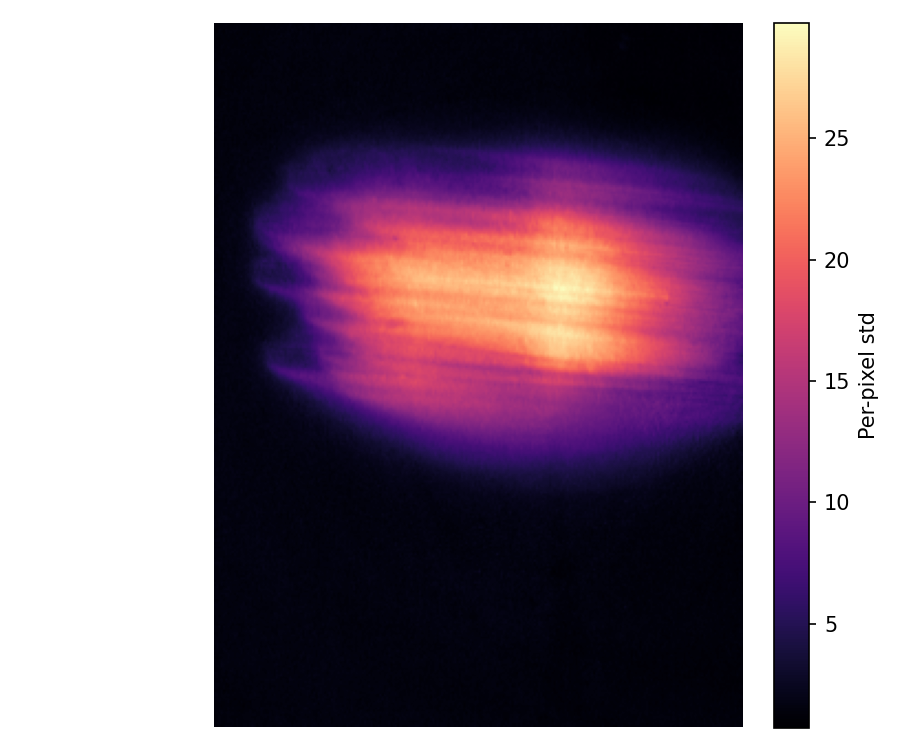}
    \caption{}
  \end{subfigure}
\end{minipage}

\caption{In-hand GelSight distributions after reset across Gear Assembly (a-b), Board Wiping (c-f), Lamp Install (e-f), Key In Lock (g-h) and Lighbulb Connect (i-j). Each column shows a $3\times3$ montage on top and the per-pixel standard deviation at the botton of the respective training dataset.}
\label{fig:appendix-reset-five-in-one-row}
\end{figure}

\subsection{Results overview}
\label{subsec:results_overview} 
Table~\ref{tab:fm_all_task_success_combined} summarizes all 1700 policy evaluation runs across both datasets (20 trials per model and task), pooling the results of models evaluated on both. Table~\ref{tab:fm_all_task_success_initial} reports policies trained on the first dataset: the main models, a sensor ablation, and an ablation of multi-tactile dropout (MTD) training. Table~\ref{tab:rebuttal_task_success_all} reports policies trained on a second, independently collected dataset: the main models, MTD inference with and without evetac observation, and additional sensor-fusion architectures (PolyTouch with cross-attention and Q-Former with learnable queries).

\begin{table*}[ht]
\centering
\caption{Policy performance across all five manipulation tasks, reported as successes/trials.
Unless marked, each entry is out of 20 trials.
$^\dagger$Evaluated on two independent datasets with successes and trials pooled, i.e., out of 40.}
\label{tab:fm_all_task_success_combined}
\renewcommand{\arraystretch}{1.1}
\setlength{\tabcolsep}{2.7pt}
\small
\begin{tabular}{llc|ccccc|cc}
\toprule
\textbf{Model} & \textbf{Sensors} &
\textbf{MTD} &
\shortstack{\textbf{Gear} \\ \textbf{Assembly}} &
\shortstack{\textbf{Board} \\\textbf{Wiping}} &
\shortstack{\textbf{Lamp} \\\textbf{Installation}} &
\shortstack{\textbf{Key} \\\textbf{in Lock}} &
\shortstack{\textbf{Lightbulb} \\ \textbf{Connection}} &
\textbf{Total} &
\textbf{Avg.} \\
\midrule
ViT-CNN$^\dagger$ & V     & --           & 15/40 & 28/40 & 13/40 & 7/40  & 6/40  & 69/200  & 34.5\% \\
ViT$^\dagger$     & V     & --           & 15/40 & 24/40 & 19/40 & 9/40  & 4/40  & 71/200  & 35.5\% \\
MiTaS      & V+E   & --           & 0/20  & 16/20 & 13/20 & 0/20  & 8/20  & 37/100  & 37\% \\
Sparsh-X     & V+G   & --           & 5/20  & 10/20 & 11/20 & 4/20  & 3/20  & 33/100  & 33\% \\
MiTaS      & V+G   & --           & 9/20  & 14/20 & 13/20 & 11/20 & 4/20  & 51/100  & 51\% \\
Sparsh-X     & V+G   & $\checkmark$ & 3/20  & 17/20 & 14/20 & 1/20  & 2/20  & 37/100  & 37\% \\
MiTaS$^\dagger$ & V+G & $\checkmark$ & 22/40 & 33/40 & 23/40 & 17/40 & 18/40 & 113/200 & 56.5\% \\
MiTaS & V+G+E & $\checkmark$ & 12/20 & 17/20 & 13/20 & 11/20 & 10/20 & 63/100  & 63\% \\
Q-Former   & V+G+E & --           & 10/20 & 15/20 & 15/20 & 12/20 & 11/20 & 63/100  & 63\% \\
Poly       & V+G+E & --           & 11/20 & 14/20 & 11/20 & 11/20 & 11/20 & 58/100  & 58\% \\
Sparsh-X$^\dagger$ & V+G+E & --       & 18/40 & 32/40 & 27/40 & 14/40 & 20/40 & 111/200 & 55.5\% \\
\textbf{MiTaS (Ours)}$^\dagger$ & V+G+E & -- & \textbf{34/40} & \textbf{38/40} & \textbf{33/40} & \textbf{30/40} & \textbf{25/40} & \textbf{160/200} & \textbf{80\%} \\
\bottomrule
\end{tabular}
\end{table*}
\begin{table*}[ht]
\centering
\caption{Policy performance of the \underline{first} dataset across all manipulation tasks (success rate in \%). Bold: best in column. Avg.: mean over the five evaluated tasks.}
\label{tab:fm_all_task_success_initial}
\renewcommand{\arraystretch}{1.15}
\setlength{\tabcolsep}{5pt}
\small
\begin{tabular}{lll|ccccc|c}
\toprule
\textbf{Model} &
\textbf{Sensors} &
\shortstack{\textbf{Co-}\\\textbf{Train}} &
\shortstack{\textbf{Gear} \\ \textbf{Assembly}} &
\shortstack{\textbf{Board} \\\textbf{Wiping}} &
\shortstack{\textbf{Lamp} \\\textbf{installation}} &
\shortstack{\textbf{Key} \\\textbf{in Lock}} &
\shortstack{\textbf{Lightbulb} \\ \textbf{Connection}} &
\textbf{Avg.} \\
\midrule
ViT-CNN    & V     & --          & 40\% & 70\% & 20\% &  0\% & 0\%  & 26\% \\
ViT        & V     & --          & 40\% & 60\% & 55\% &  0\% & 0\%  & 31\% \\
MiTaS      & V+E   & --          &  0\% & 80\% & 65\% &  0\% & 40\%  & 37\% \\
Sparsh     & V+G   & --          & 25\% & 50\% & 55\% & 20\% & 15\%  & 33\% \\
MiTaS      & V+G   & --          & 45\% & 70\% & 65\% & 55\% & 20\%  & 51\% \\
Sparsh     & V+G   & $\checkmark$ &  15\% & 85\% & 70\% &  5\% & 10\%  & 37\% \\
MiTaS      & V+G   & $\checkmark$ & 55\% & 85\% & 60\% & 20\% & 45\%  & 53\% \\
Sparsh     & V+G+E & --          & 50\% & 85\% & 70\% & 10\% & 55\%  & 54\% \\
MiTaS (Ours) & V+G+E & -- & \textbf{90\%} & \textbf{90\%} & \textbf{80\%} & \textbf{75\%} & \textbf{65\%}  & \textbf{80\%} \\
\bottomrule
\end{tabular}
\vspace{2pt}
\end{table*}

\begin{table*}[ht]
\centering
\caption{Policy performance of the \underline{second} dataset across all manipulation tasks (success rate in \%, 20 trials per task). Bold: best in column. Avg.: mean over the five tasks}
\label{tab:rebuttal_task_success_all}
\renewcommand{\arraystretch}{1.15}
\setlength{\tabcolsep}{4.5pt}
\small
\begin{tabular}{llc|ccccc|c}
\toprule
\textbf{Model} & \textbf{Sensors} &
\textbf{MTD} &
\shortstack{\textbf{Gear} \\ \textbf{Assembly}} &
\shortstack{\textbf{Board} \\\textbf{Wiping}} &
\shortstack{\textbf{Lamp} \\\textbf{Installation}} &
\shortstack{\textbf{Key} \\\textbf{in Lock}} &
\shortstack{\textbf{Lightbulb} \\ \textbf{Connection}} &
\textbf{Avg.} \\
\midrule
ViT-CNN    & V     & --           & 35\% & 70\%  & 45\% & 35\% & 30\% & 43\% \\
ViT        & V     & --           & 35\% & 60\%  & 40\% & 45\% & 20\% & 40\% \\
MiTaS       & V+G   & $\checkmark$ & 55\% & 80\%  & 55\% & 65\% & 45\% & 60\% \\
MiTaS       & V+G+E & $\checkmark$ & 60\% & 85\%  & 65\% & 55\% & 50\% & 63\% \\
Q-Former   & V+G+E & --           & 50\% & 75\%  & 75\% & 60\% & 55\% & 63\% \\
Poly       & V+G+E & --           & 55\% & 70\%  & 55\% & 55\% & 55\% & 58\% \\
Sparsh-X   & V+G+E & --           & 40\% & 75\%  & 65\% & 60\% & 45\% & 57\% \\
\textbf{MiTaS (Ours)} & V+G+E & -- & \textbf{80\%} & \textbf{100\%} & \textbf{85\%} & \textbf{75\%} & \textbf{60\%} & \textbf{80\%} \\
\bottomrule
\end{tabular}
\end{table*}






~ \clearpage
\subsection{Attention analysis}
\label{subsec:attentionanalysis_appen}

\begin{figure}[ht!]
  \centering
  \begin{subfigure}{\linewidth}
    \centering
    \includegraphics[width=0.67\columnwidth]{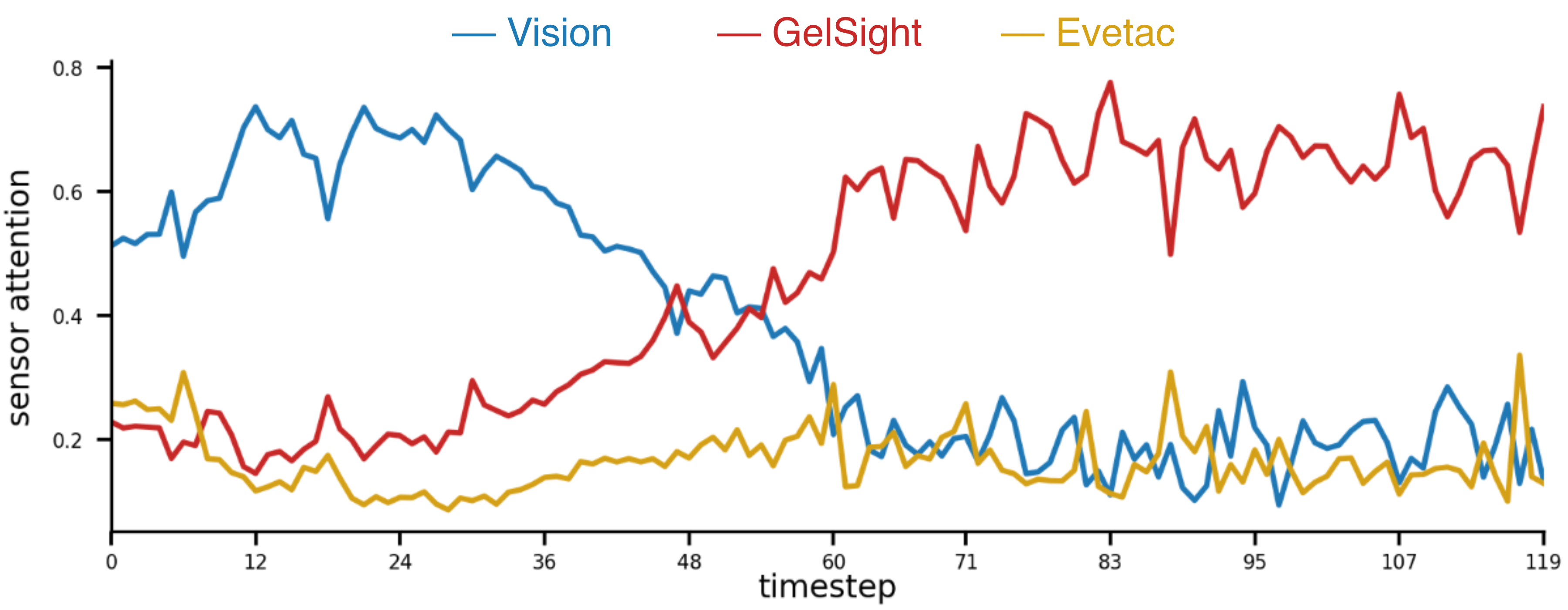}
    \caption{Gear Assembly}
    \label{fig:attentionplots:a}
  \end{subfigure}
  \vspace{0.4em}
  \begin{subfigure}{\linewidth}
    \centering
    \includegraphics[width=0.67\columnwidth]{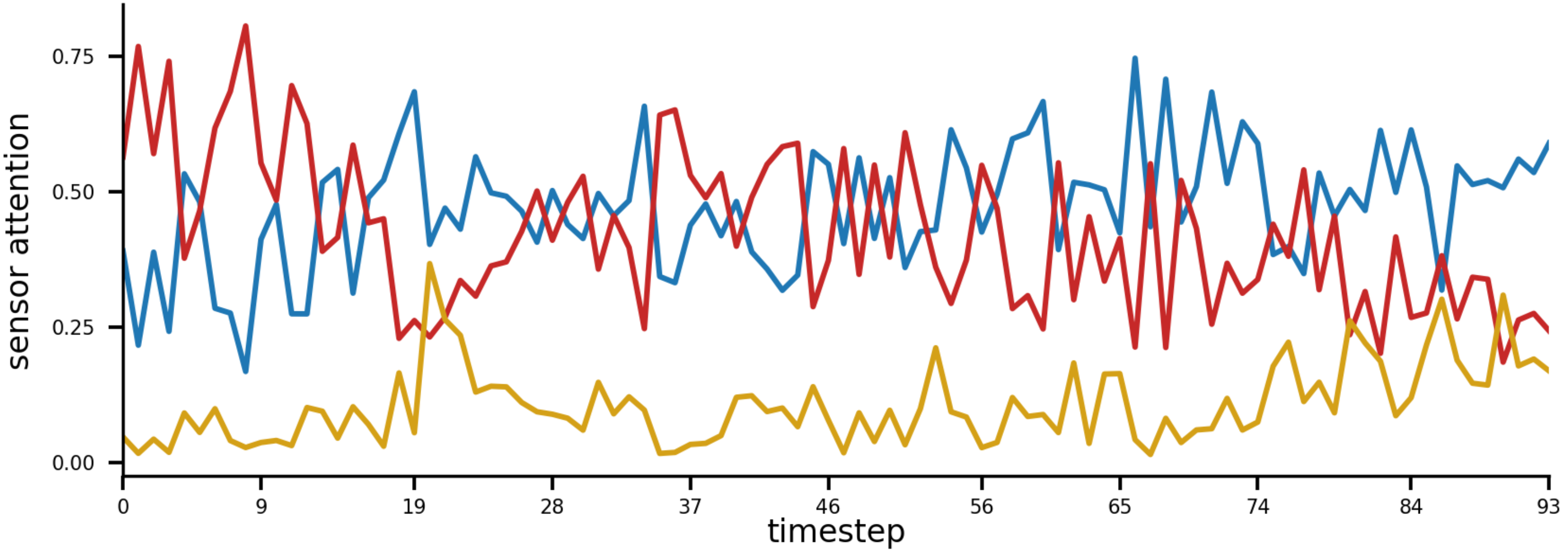}
    \caption{Board Wiping}
    \label{fig:attentionplots:b}
  \end{subfigure}
  \vspace{0.4em}
  \begin{subfigure}{\linewidth}
    \centering
    \includegraphics[width=0.67\columnwidth]{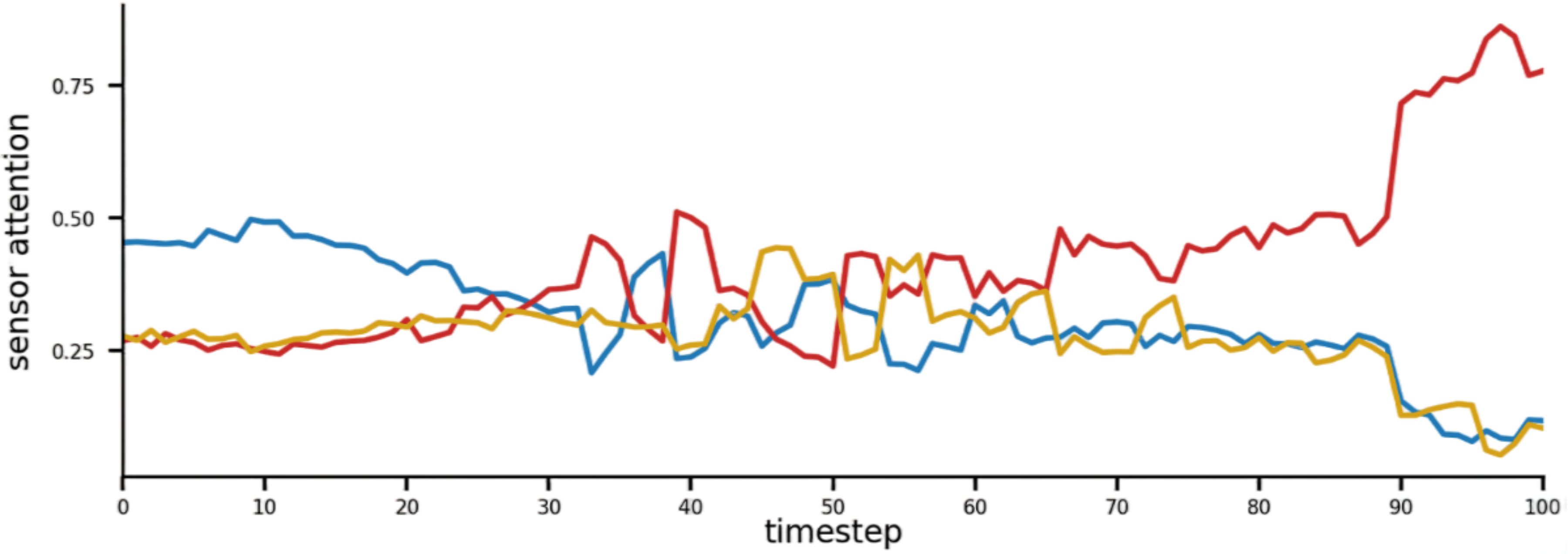}
    \caption{Lamp Installation}
    \label{fig:attentionplots:d}
  \end{subfigure}
  \vspace{0.4em}
  \begin{subfigure}{\linewidth}
    \centering
    \includegraphics[width=0.67\columnwidth]{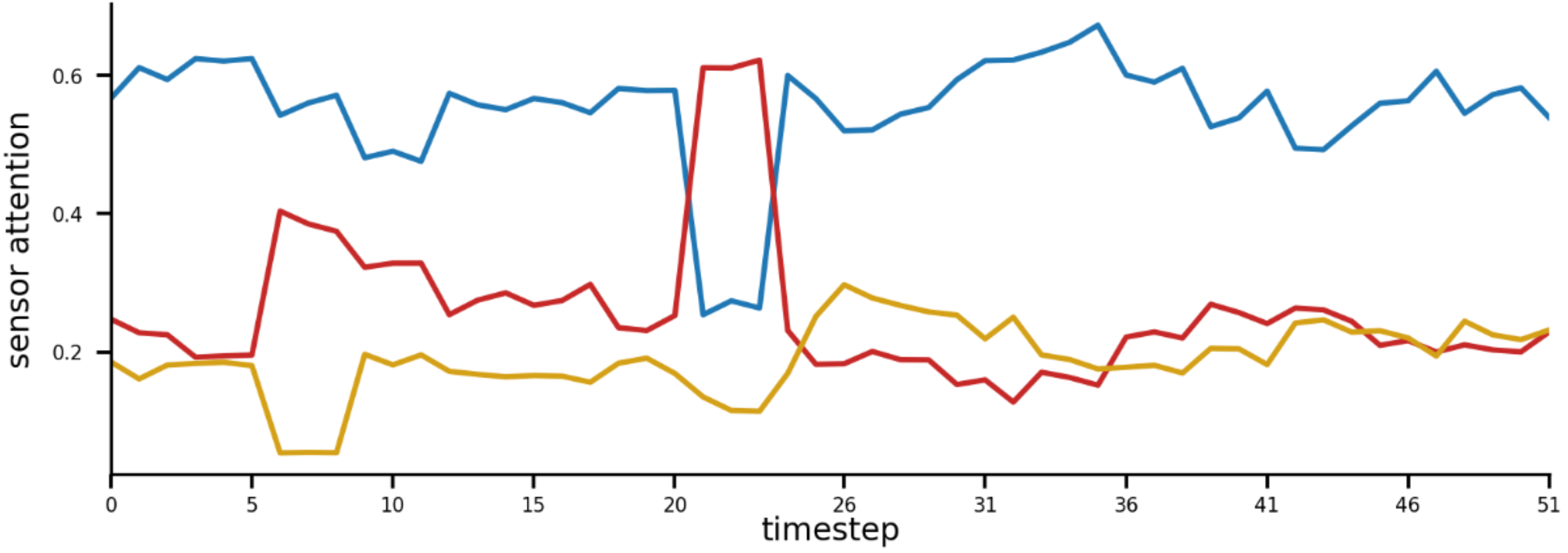}
    \caption{Key in Lock}
    \label{fig:attentionplots:c}
  \end{subfigure}
  \vspace{0.4em}
  \begin{subfigure}{\linewidth}
    \centering
    \includegraphics[width=0.67\columnwidth]{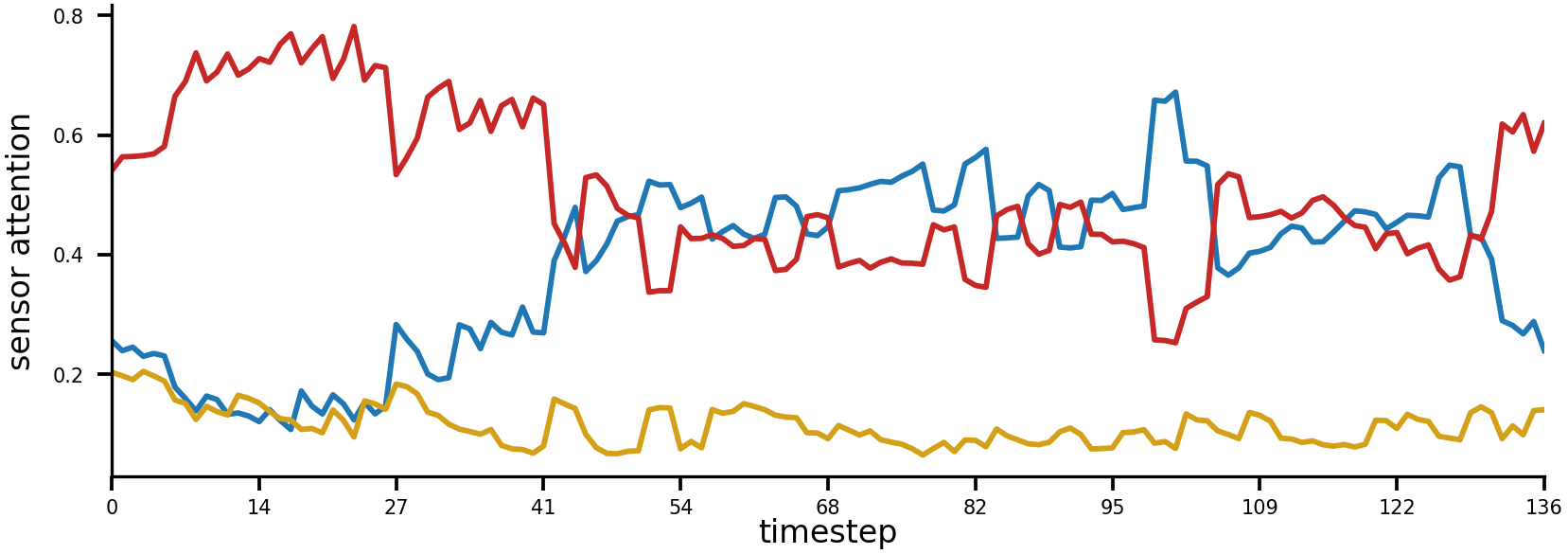}
    \caption{Lightbulb Connection}
    \label{fig:attentionplots:e}
  \end{subfigure}
  \caption{For each task we plot the cross-attention from the executed action token to each sensor modality (Vision, GelSight, Evetac) over the course of a successful episode.}
  \label{fig:attention_plots}
\end{figure}

\newpage
\subsection{Evetac Activation}

\begin{figure}[ht!]
  \centering
  \begin{subfigure}{\linewidth}
    \centering
    \includegraphics[width=0.7\columnwidth]{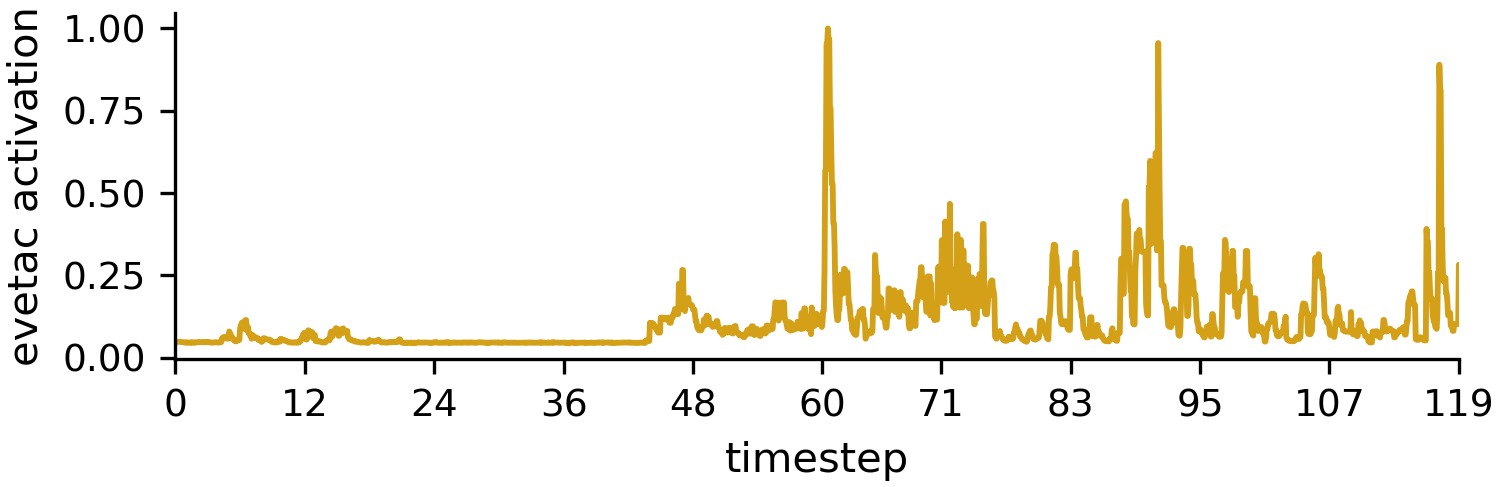}
    \caption{Gear Assembly}
    \label{fig:stack:a}
  \end{subfigure}
  \vspace{0.4em}
  \begin{subfigure}{\linewidth}
    \centering
    \includegraphics[width=0.7\columnwidth]{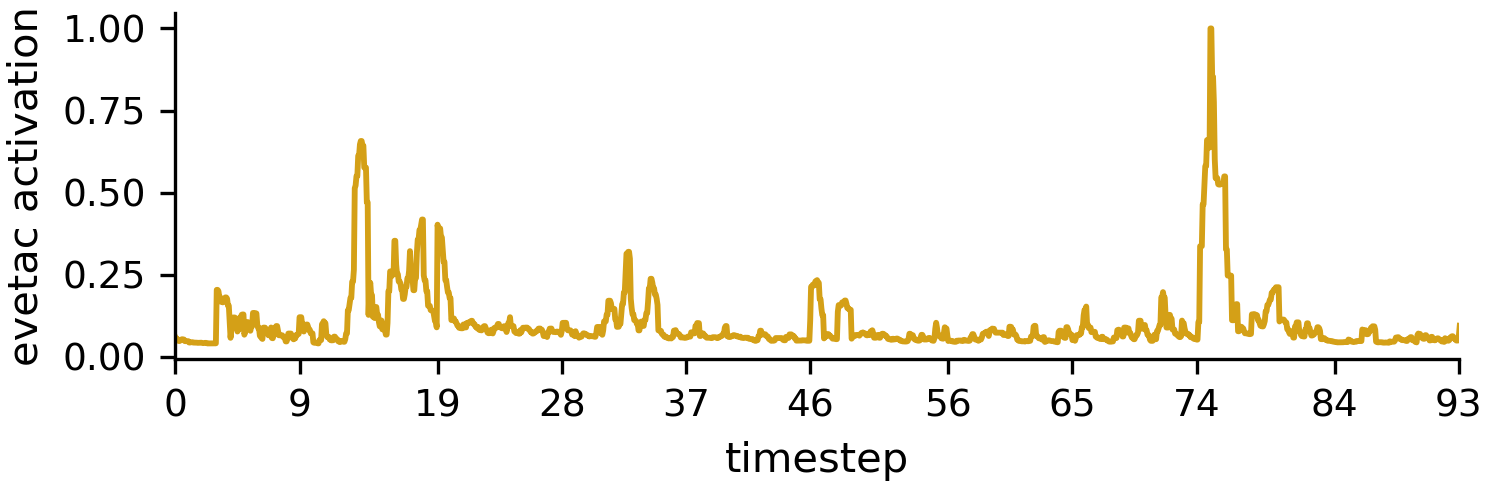}
    \caption{Board Wiping}
    \label{fig:stack:b}
  \end{subfigure}
  \vspace{0.4em}
  \begin{subfigure}{\linewidth}
    \centering
    \includegraphics[width=0.7\columnwidth]{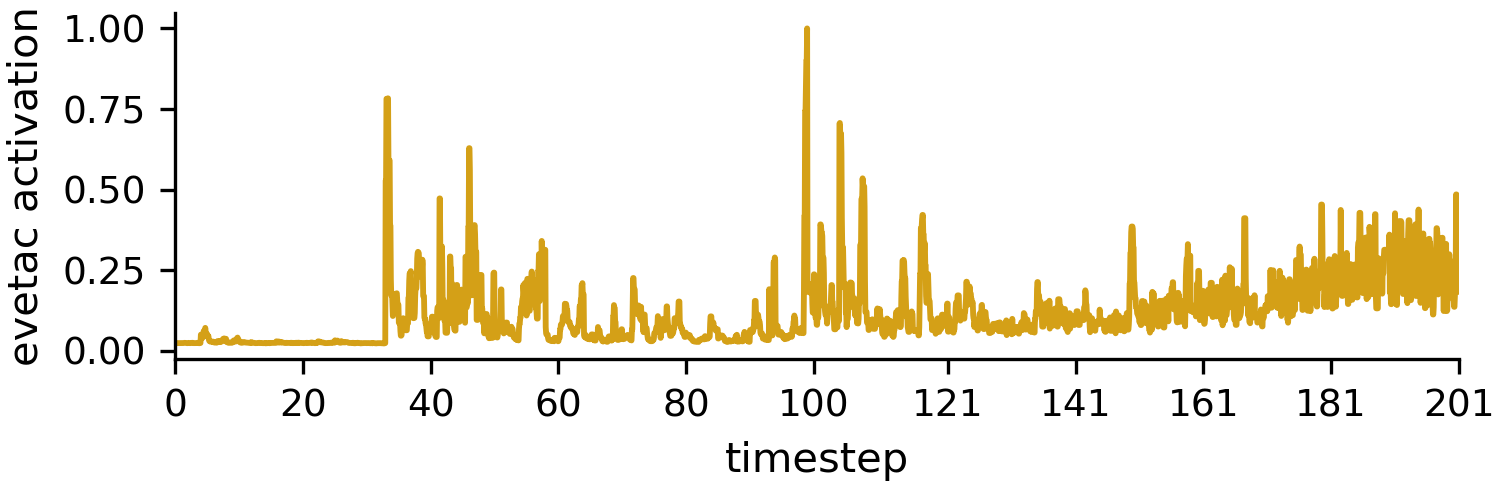}
    \caption{Lamp Installation}
    \label{fig:stack:d}
  \end{subfigure}
  \vspace{0.4em}
  \begin{subfigure}{\linewidth}
    \centering
    \includegraphics[width=0.7\columnwidth]{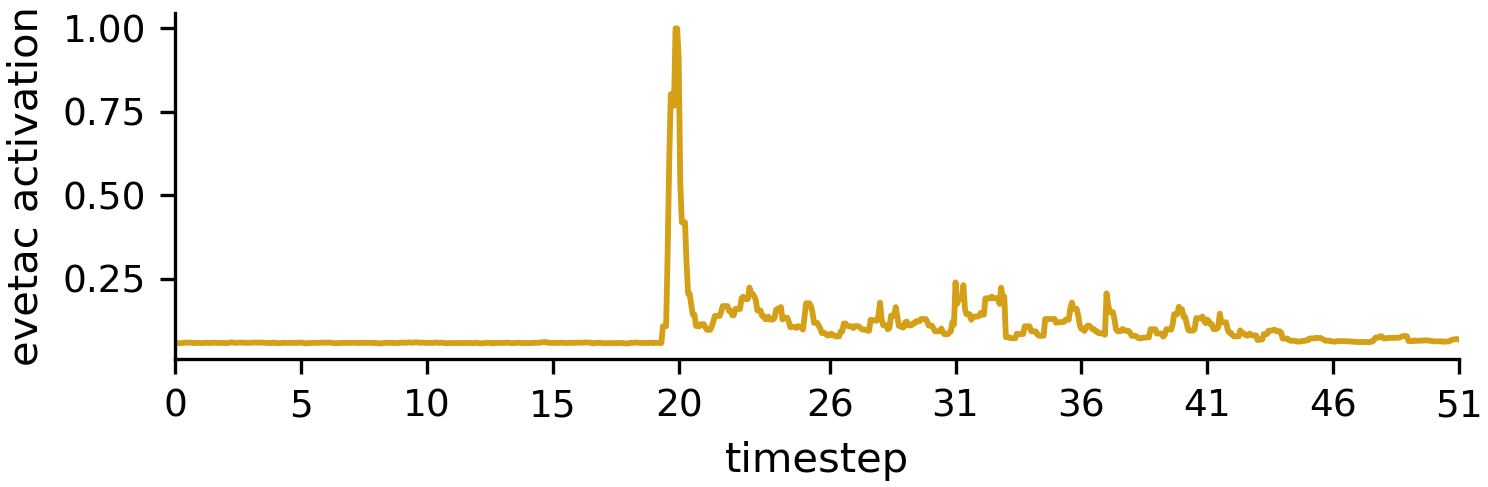}
    \caption{Key in Lock}
    \label{fig:stack:c}
  \end{subfigure}
  \vspace{0.4em}
  \begin{subfigure}{\linewidth}
    \centering
    \includegraphics[width=0.7\columnwidth]{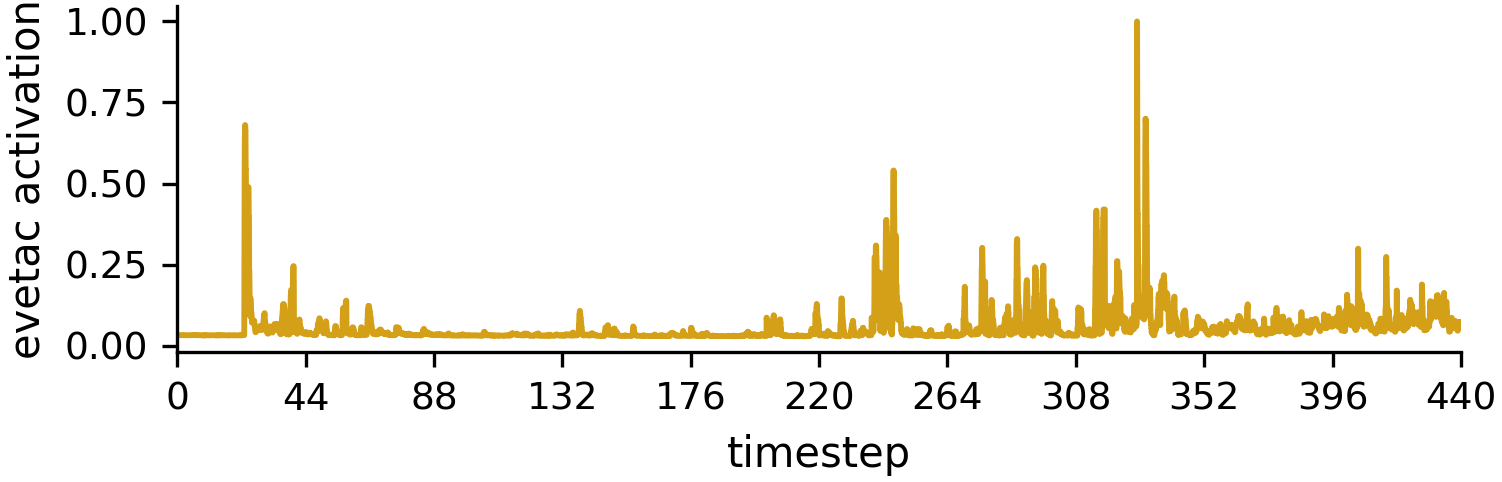}
    \caption{Lightbulb Connection}
    \label{fig:stack:e}
  \end{subfigure}
  \caption{We define the Evetac activation as the L1-distance to the base image in pixel-space. Evetac activations differ for each task and task-stage indicating its usefullness to the policy. During reaching, beginning of a trajectory, the activation is low, it spikes by contact and provides high frequency features during object interaction.}
  \label{fig:stack}
\end{figure}

\end{document}